# Flight Delay Prediction using Hybrid Machine Learning Approach: A Case Study of Major Airlines in the United States

Rajesh Kumar Jha[1], Shashi Bhushan Jha[2,4,*], Vijay Pandey[3], Radu F. Babiceanu[4]

[1]Department of Electronics and Communication Engineering, BNMIT, India
[2]Department of Computer Science, University of West Florida, FL, USA
[3]Department of Computer Science Engineering, IIT Kharagpur, India
[4]Department of Electrical Engineering and Computer Science, Embry-Riddle Aeronautical University, Daytona Beach, USA

E-mail: sjha@uwf.edu, babicear@erau.edu, vijayiitkgp13@gmail.com, rajeshjnv23@gmail.com

[*]Corresponding author (email: sjha@uwf.edu)

**Abstract**

The aviation industry has experienced constant growth in air traffic since the deregulation of the U.S. airline industry in 1978. As a result, flight delays have become a major concern for airlines and passengers, leading to significant research on factors affecting flight delays such as departure, arrival, and total delays. Flight delays result in increased consumption of limited resources such as fuel, labor, and capital, and are expected to increase in the coming decades. To address the flight delay problem, this research proposes a hybrid approach that combines the feature of deep learning and classic machine learning techniques. In addition, several machine learning algorithms are applied on flight data to validate the results of proposed model. To measure the performance of the model, accuracy, precision, recall, and F1-score are calculated, and ROC and AUC curves are generated. The study also includes an extensive analysis of the flight data and each model to obtain insightful results for U.S. airlines.

## I. Introduction

Over the last few decades, studying the air transportation system has become a crucial area of research, especially in relation to flight delays caused by high demand and limited capacity. Since the deregulation of the U.S. airline industry in 1978, the aviation industry has experienced constant growth in air traffic, resulting in increased competition and a significant amount of research on factors affecting flight delays such as departure, arrival, and total delays. This growth has also led to a global increase in air traffic, resulting in numerous delays and costs for both airlines and passengers (World Bank, 2013) (Ball et al., 2010; JEC, 2008; Cook et al., 2004). The Federal Aviation Administration (FAA) predicts a 40% increase in total enplanements by 2038, with increased operating costs for airlines being a major concern due to flight delays (Zou and Chen, 2017). Such delays result in increased consumption of fuel, labor, capital, and other limited resources. According to the FAA, delay and congestion are expected to increase in the coming decades, which will further exacerbate these issues (Boeing, 2011).

Initially, the airline industry required information on actual flight departure and arrival times, as well as details on flight cancellations and diversions. Between 1995 and 2008, the aviation industry was also required to disclose several factors related to flight delays, including the cause of delays, flight cancellations, technical issues, airborne times, aircraft tail numbers, and taxi times. Moreover, the tarmac rule stipulates that carriers are not permitted to keep passengers on board for more than three hours without deplaning them (Yimga and Gorjidooz, 2019; Forbes et al., 2019).

This research aims to address the flight delay problem by categorizing it into three distinct subproblems - flight departure delay, flight arrival delay, and total flight delay - and using important features of flight operations. To solve these problems, the study develops a hybrid approach that combines deep learning and classic machine learning techniques. Additionally, various machine learning algorithms are used to validate the results of the proposed approach. The performance of all flight delay classification problems is measured based on accuracy, precision, recall, and F1-score, and ROC and AUC curves are generated. This study also includes an extensive analysis of each model to obtain insightful results for U.S. airlines.

The organization of the paper is structured in the following way: In Section 2, the literature on the flight delay problem is reviewed. Section 3 examines the flight delay problem, analysing the flight dataset and presenting various plots to showcase the key insights. Section 4 introduces the proposed hybrid approach and other machine learning techniques for addressing the problem. Section 5 illustrates the empirical results of the flight delay problem. Section 6 provides a discussion of the results, and in Section 7, the conclusion and future scope of this study is presented.

## II. Literature Review

Previous research has addressed various significant factors related to flight delay. Zou et al. (2014) conducted an extensive empirical analysis to examine the impact of flight delay on airfare and flight frequency of U.S. airlines. The authors modeled airfare and flight frequency as function costs and considered the demand characteristics and competition effects. The authors collected the dataset from the FAA and the Bureau of Transportation Statistics. They divided the problem into two sections: the first section involved econometric models with non-stop routes that considered 13 features, while the second section involved econometric models with one-stop routes that incorporated 17 features. The authors used a regression model to provide a more comprehensive understanding of the impact of delay on airline pricing and frequency scheduling. The results showed that transportation agencies tend to pass the cost of delay on to passengers through higher fares. Lambelho et al. (2020) presented a comprehensive evaluation of strategic flight schedules using prediction techniques for flight delays and cancellations. The research focused on flight schedules from 2013 to 2018 at London Heathrow Airport and utilized the dataset to showcase their classification model. The authors utilized the recursive feature elimination (RFE) method to select crucial features. The study developed a machine learning approach to assess flight schedules by predicting flight delays and cancellations. This approach was aimed at supporting the slot allocation process at airports. Yu et al. (2019) analysed a high-dimensional dataset from Beijing International Airport to establish an effective model for predicting flight delays. They developed a novel approach that combined the deep belief network with the support vector regression model to create a supervised fine-tuning predictive architecture. This model demonstrated excellent performance in handling large datasets and effectively capturing the primary factors that contribute to delays.

Yimga (2020) conducted a study on the influence of flight delay on market power in the US airline industry. The study focused on estimating the effect of flight delays on product markups and utilized a structural econometric model of demand and supply to break down the estimated markup effects into price and marginal cost effects. The findings revealed that flight delays have adverse effects on consumer welfare. Rebollo and Balakrishnan (2014) introduced a new category of models that can predict air traffic delays and incorporate both temporal and spatial delay factors. This research employed the Random Forest algorithm to forecast departure delays and evaluated the performance of the proposed prediction models in classifying delays and predicting delay values. The authors used operational data from 2007 and 2008 and analyzed the 100 most delayed links of U.S. airports. The outcomes showed that, for a two-hour forecast horizon, the mean test error over the 100 links was 19% when classifying delays as above or below 60 minutes. Liu et al. (2019) investigated the influence of weather and arrival delay on the incidence of ground delay program (GDP) and used Support Vector Machine (SVM) to develop a prediction model. The study analyzed how regional convective weather impacts GDP incidence and discovered that the effect depends on the direction and distance of convective activity from the airport. The SVM-generated regional weather variables, local weather, and arrival demand were used to train the model, and its performance was compared with logistic regression and random forest. The outcomes demonstrated that random forest produced better results than logistic regression, and the researchers concluded that convective weather is the most critical factor in predicting GDP incidence at Atlanta International Airport.

Although the numerous publications on the issue of flight delays, there is a scarcity of literature pertaining to the prediction of flight delays specifically for U.S. airlines. Furthermore, there is a need to enhance the quality of the existing solutions. This study makes a significant contribution in five distinct areas:

1) Contributes to the review of existing literature.
2) Collects 27 months of flight data of U.S. airlines considering various factors.
3) Conducts data analysis on the flight dataset.
4) Develops a hybrid technique that amalgamates the attributes of both deep learning and traditional machine learning models for U.S. airlines flight dataset. To the authors' knowledge, the developed method has not yet been explored to predict the flight delay for U.S. airlines in past studies.

## III. Problem Description

This section presents an overview of the problem and data analysis conducted on the U.S. airlines dataset. The study analyses a 27-month flight dataset of 11 market and 28 operating U.S. airlines between January 2018 and March 2020. The dataset is obtained from the Bureau of Transportation Statistics, United States Department of Transportation[1]. The primary objective of the study is to forecast flight arrival, departure, and total delay using a novel hybrid approach and validated with various machine learning techniques. Initially, the dataset contains 17.93 million records classified by marketing carrier airlines, as shown in Table 1. Most flights are operated by a few U.S. airlines, including American Airlines, Delta Airlines, United Airlines, and Southwest Airlines. The dataset consists of 119 features; however, several features are over 90% missing values and are therefore excluded. Subsequently, the dataset is reduced to 18 million records with 70 features. Due to the large number of records, two thousand records are randomly selected from each month over the 27-month period, and non-essential and repetitive features are excluded, resulting in a total of 54,000 records with 37 features. Out of these records, 40,500 are used for training, and the remaining 13,500 are reserved for testing purposes.

[1] https://www.transtats.bts.gov/DataIndex.asp

**Table 1. Number of flights categorized with U.S. marketing airlines.**

| U.S. Marketing Airlines | Marketing carrier IATA code | Number of flights |
|---|---|---|
| American Airlines | AA | 4,661,202 |
| Delta Airlines | DL | 3,920,936 |
| United Airlines | UA | 3,485,944 |
| Southwest Airlines | WN | 3,044,538 |
| Alaska Airlines | AS | 960,276 |
| JetBlue Airways | B6 | 676,314 |
| Spirit Airlines | NK | 435,593 |
| Front Airlines | F9 | 292,359 |
| Allegiant Air | G4 | 229,768 |
| Hawaiian Airlines | HA | 208,064 |
| Virgin Airlines | VX | 17,670 |

The majority of flight delay data analysis is presented through bar plots, pie charts, and line plots, based on randomly selected two thousand records per month. Both operating and marketing airlines are included in the flight dataset, so the plots are generated for both categories. Fig. 1 and 2 display the plots generated for flight delay over operating carriers. Fig. 1 illustrates the number of delayed and non-delayed flights and the frequency of flights for each carrier, while Fig. 2 shows the probability of delayed and non-delayed flights for each carrier. Southwest Airlines has the highest number of delayed flights, but also has the most non-delayed flights. Cape Air, Empire Airlines, Penair, and Virgin Airlines have a lesser number of delayed and non-delayed flights. In terms of the probability of delayed and non-delayed flights, Cape Air has the highest probability of delay, while Penair has the least probability of delay. The study also conducts data analysis of flight delay with respect to marketing airlines, as shown in Fig. 3 and 4. American Airlines has the highest number of delayed flights, but also has the most total flights as it operates eleven carriers. Delta Airlines exhibits the highest probability of delay when considering the total number of flights, while Southwest Airlines exhibits the least probability of delay.

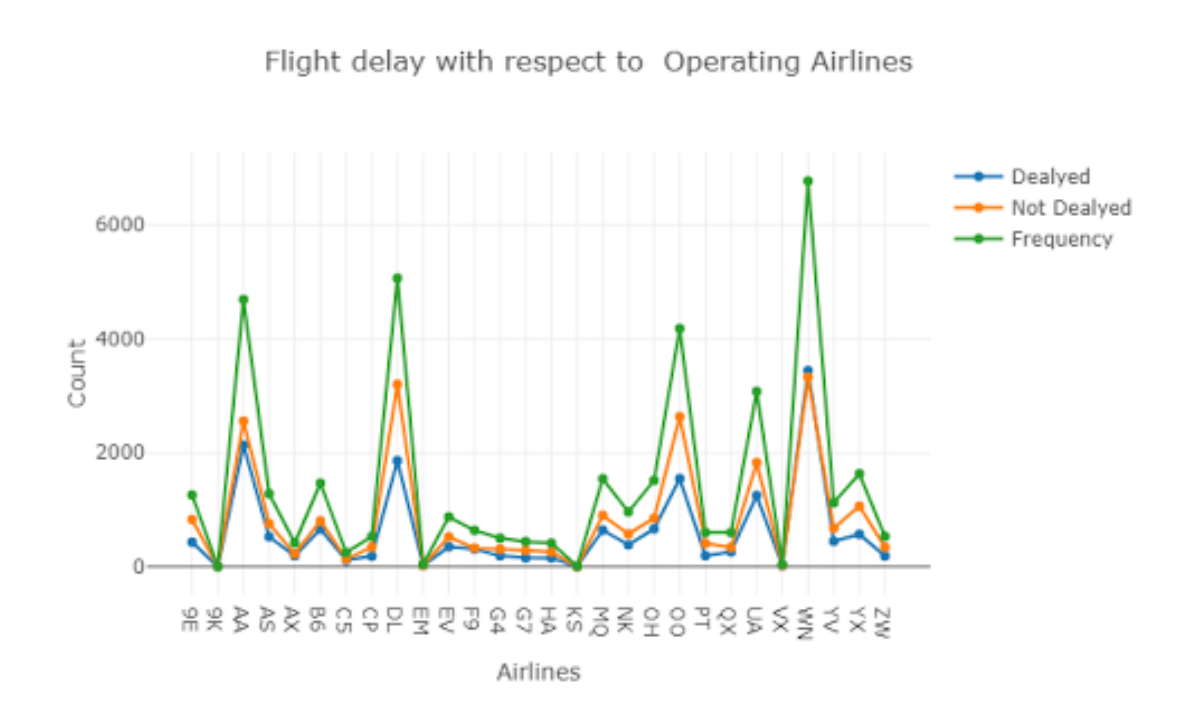

Fig. 1. The number of flights delayed with respect to operating airlines

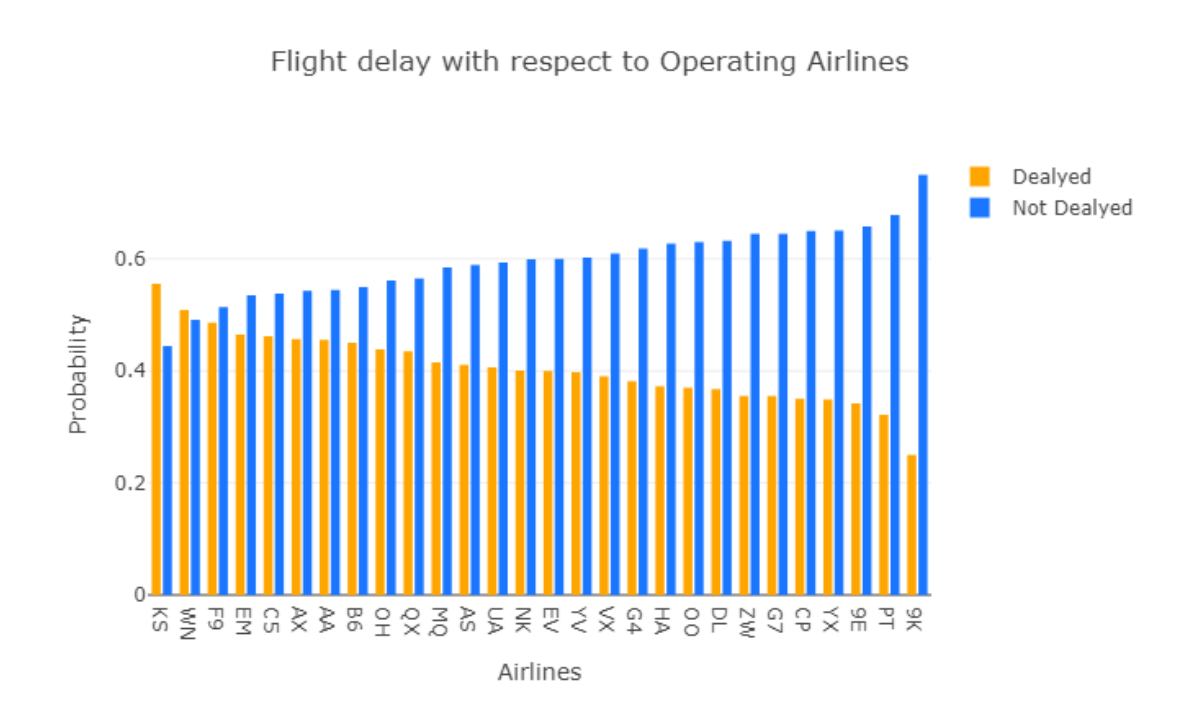

Fig. 2. The probability of flight delayed with respect to operating airlines

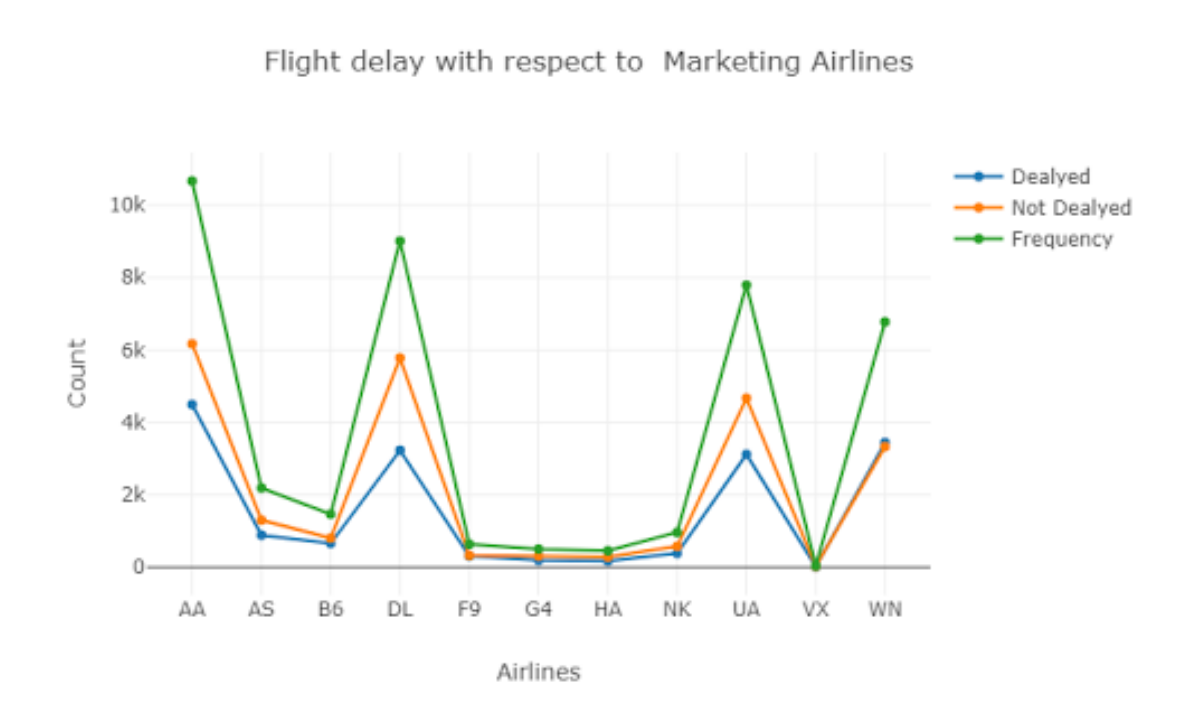

Fig. 3. The number of flights delayed over marketing airlines

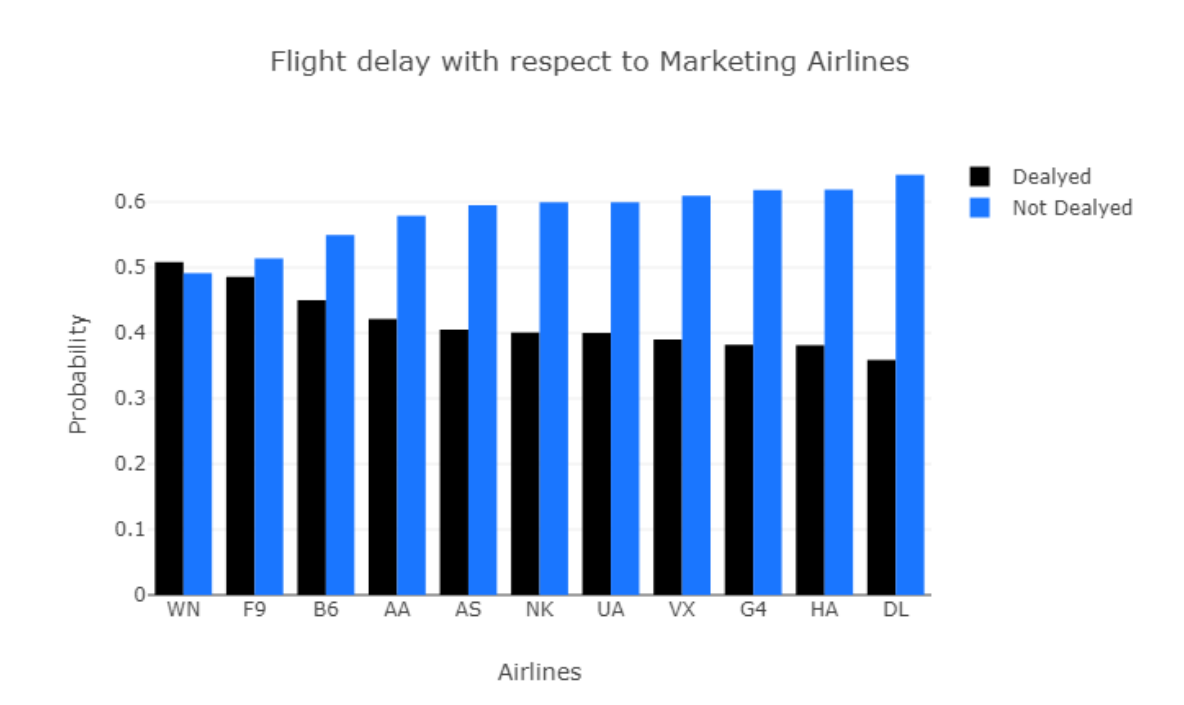

Fig. 4. The probability of flight delayed with respect to marketing airlines

Fig. 5 and 6 illustrate the flight delay plots of U.S. airlines across various U.S. States and Territories. Fig. 5 indicates that California, Texas, Georgia, Florida, and New York have the highest number of both delayed and non-delayed flights. This is likely due to the high population and economy in these states, which results in a greater frequency of flights and subsequently, more delayed and non-delayed flights. Furthermore, Fig. 6 displays the probability of flight delays in relation to each U.S. State and territory. The ratio of delayed flights to the total number of flights flown from a particular state is demonstrated. New Hampshire has the highest probability of on-time flights, while Puerto Rico has the lowest.

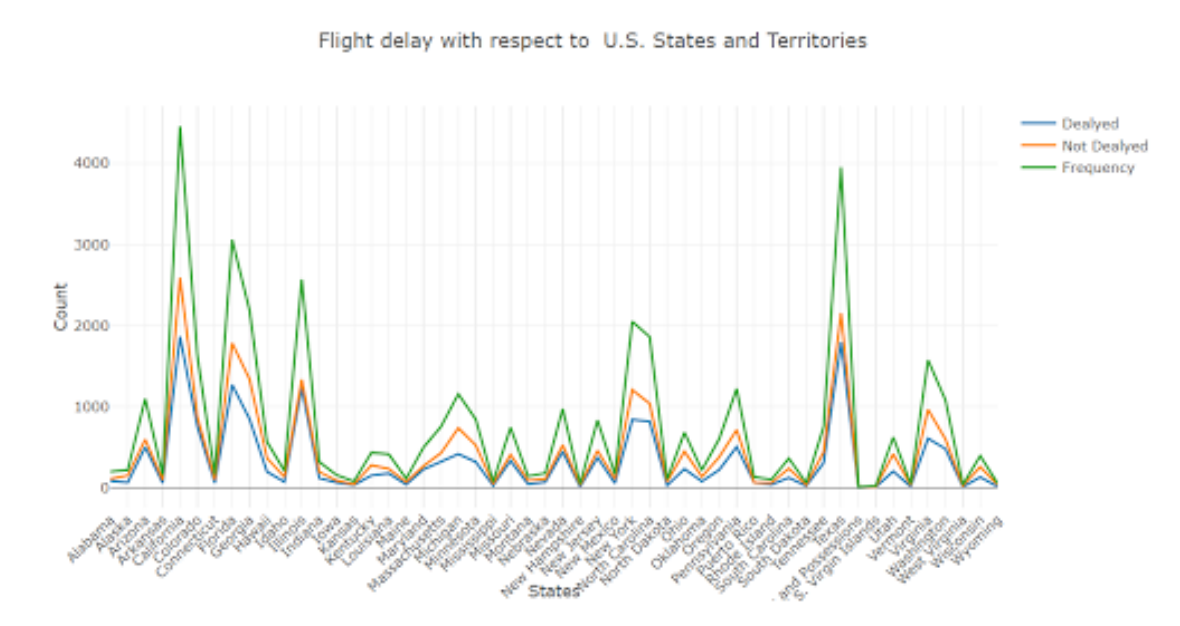

Fig. 5. The number of flights delayed with respect to the U. S. States and territories

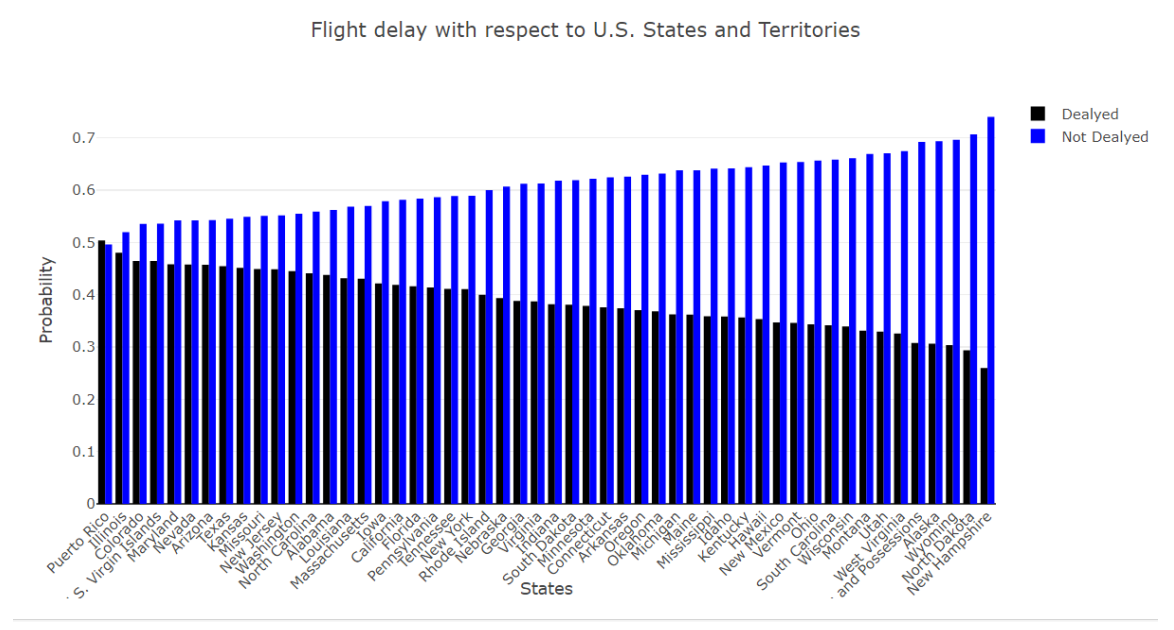

Fig. 6. The probability of flight delayed over the U. S. States and territories

Fig. 7 is generated for the flights by distance (in miles), where flight distances are categorised in short haul, medium haul, and long haul. Short haul range is between 0 to 800 miles, medium haul range interval is 800 to 2200 miles, and greater than 2200 miles distance falls under long haul. In this figure, it can be noticed that most of the U.S. airlines' flights fall under short haul and the least number of flights are categorized as long haul. In addition, one of the features is a distance group in the flight dataset that categorizes the flights based on the distance intervals for every 250 Miles. So, if the distance is between 0 to 250 miles, then it classifies it as class 0, and similarly, the same approach is followed until class 10. The pie-chart of flights over the distance groups is depicted in Fig. 8. It can be noted that as the class interval increases, the number of the flights decreases. Moreover, flights delayed and not delayed with respect to distance groups are represented in Fig. 9. The plot shows that as the class interval of the distance group grows, the probability of flight delayed lessens in almost all intervals except the last two distance groups.

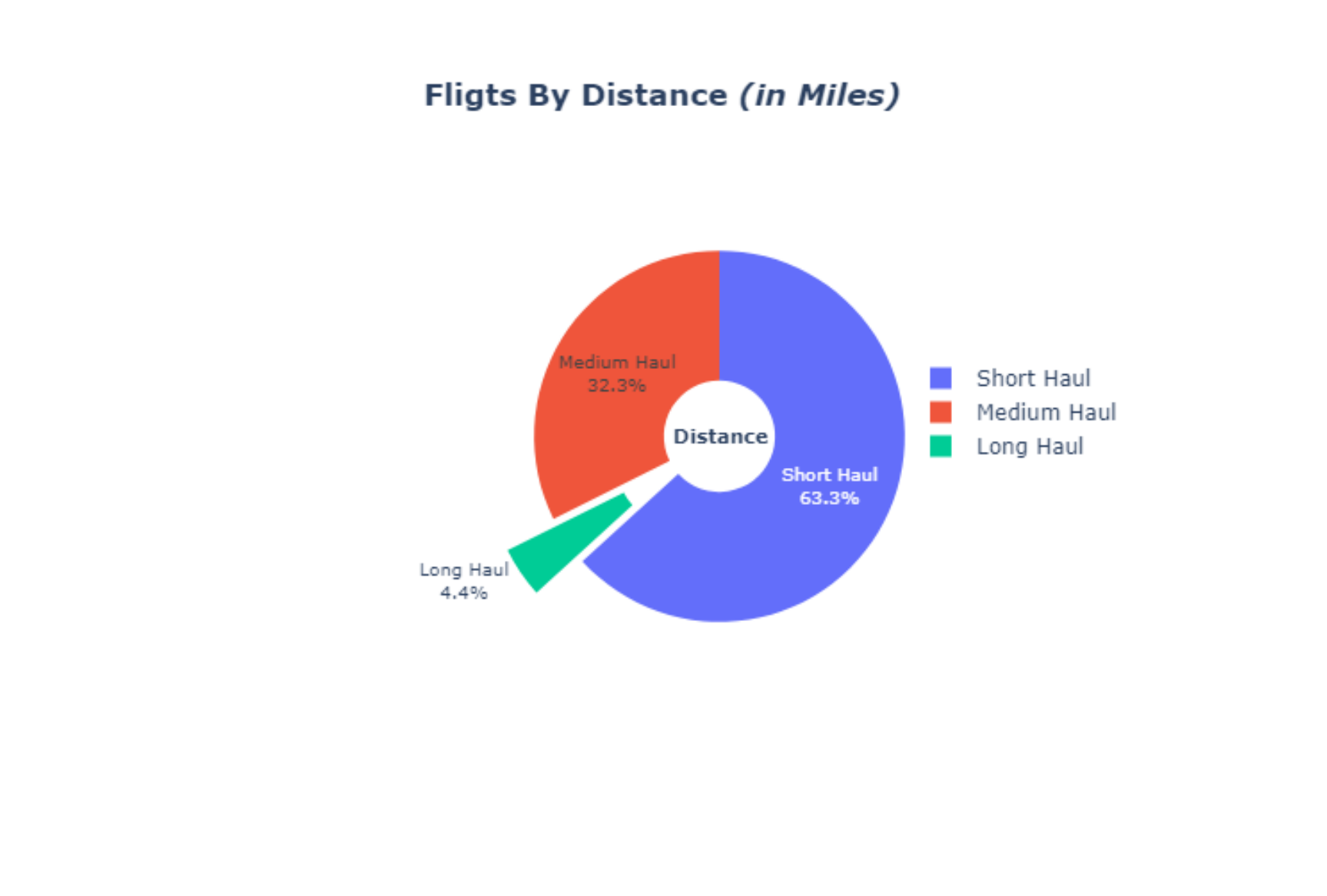

Fig. 7. The percentage of flights by distance (in miles)

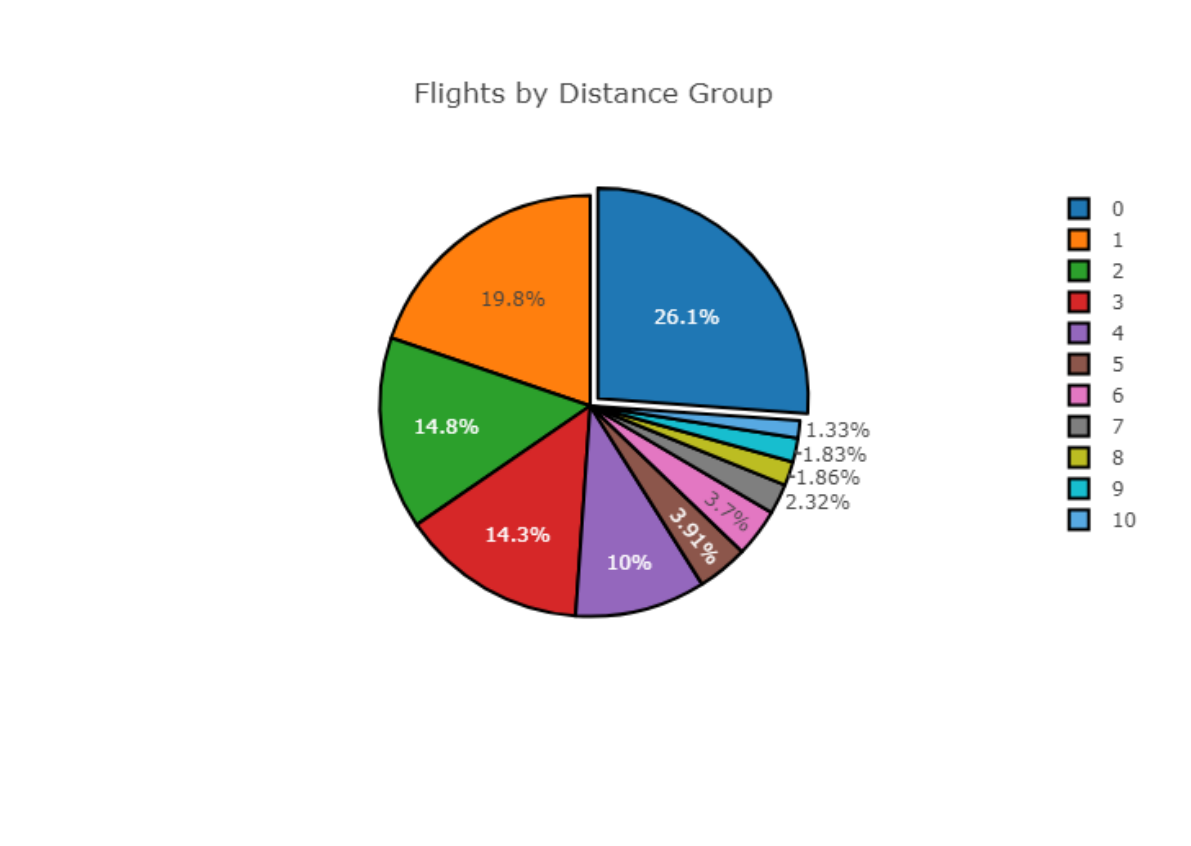

Fig. 8. The percentage of flights by distance (in miles) categorised in eleven groups

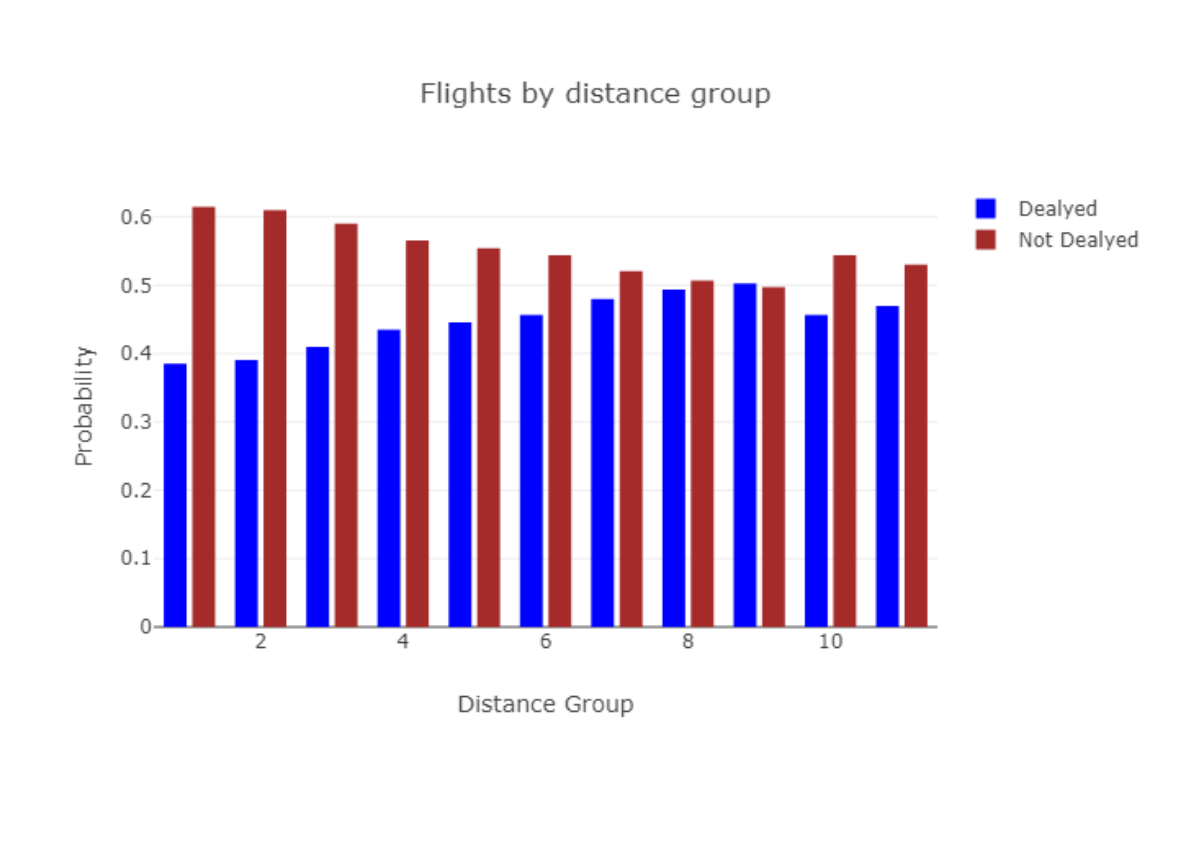

Fig. 9. The probability of flight delayed with respect to distance group

Fig. 10 and 11 analyzes the fluctuation of flight delays versus non-delays across the months. Figure 10 displays the distribution of flight delays throughout the months, revealing a concentration of delays during peak travel seasons, such as summer vacations and festive periods. Conversely, Fig. 11 illustrates the probability of flight delays for each month, calculated as the ratio of delayed flights to total flights. The month of September records the highest probability of delays, while the month of June exhibits the lowest. The decreased likelihood of delays in June is attributed to airlines anticipating traffic surges as summer vacations commence.

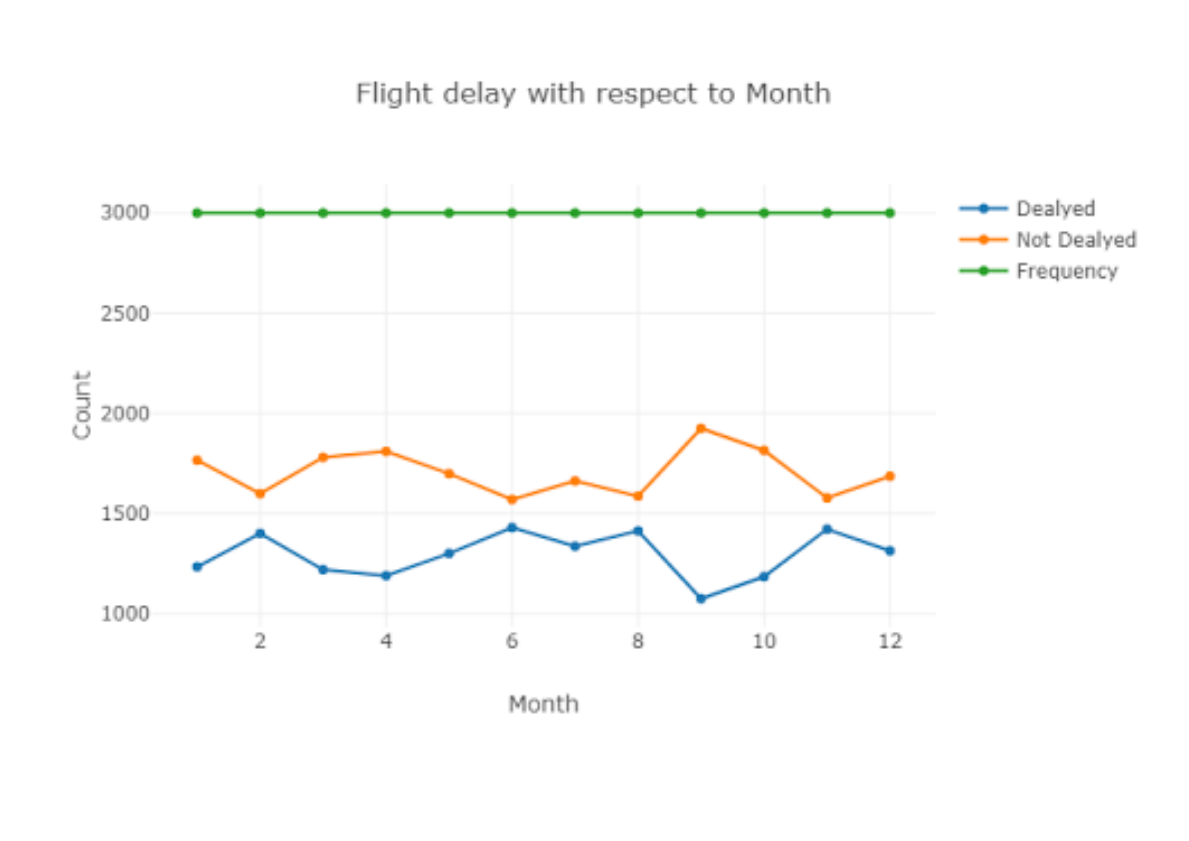

Fig. 10. The number of flights delayed over months

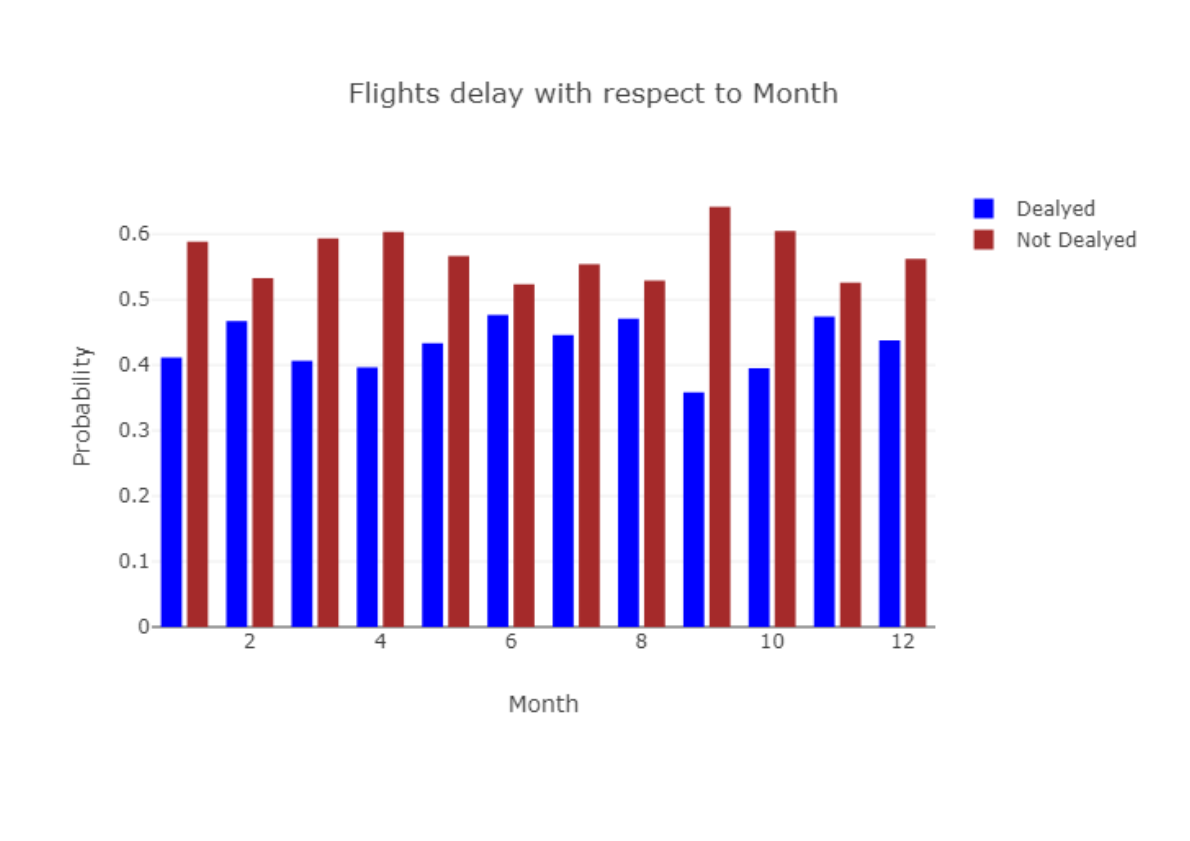

Fig. 11. The probability of flight delayed with respect to months

Fig. 12 and 13 show the plot with respect to the number of flights and probability of flight delays over days of the week, respectively. In Fig. 12, one can note that most of the flight delays occur on Friday because most of the people leave for the weekend, either travel to meet the family or to the weekend destination for leisure time. On the contrary, the least flight delays exist on Saturday since people spend spare time at a location. Furthermore, Fig. 13 exhibits that the highest and smallest probability of flight delays exist on Tuesday and Friday, respectively. The least probability of flight delay is recorded on Friday because the airlines are prepared for high frequency since they are aware of passengers' movement. However, the greatest probability of flight delay is recorded on Tuesday because airlines expect less movement of passengers on that day of the week.

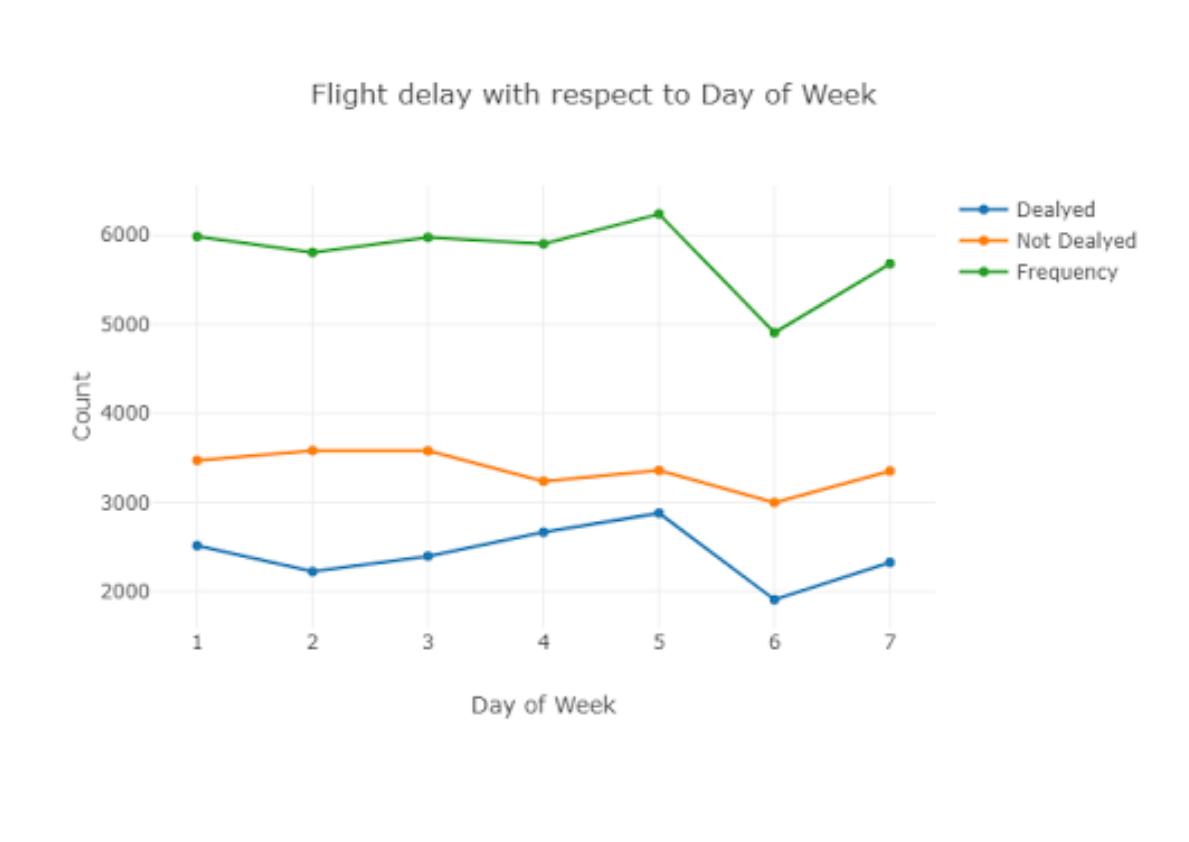

Fig. 12. The number of flights delayed with respect to the day of week

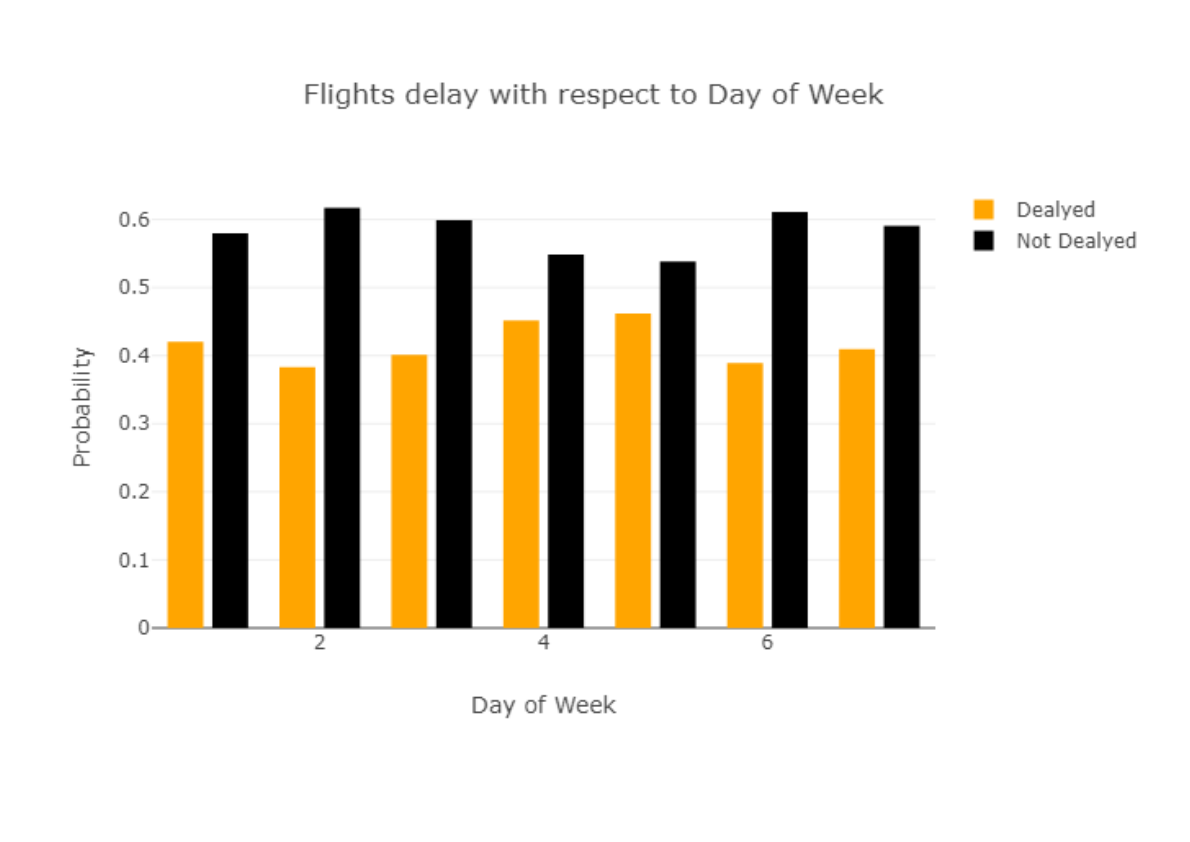

Fig. 13. The probability of flight delayed over the day of week

Another analysis of the flight data is on flight delay over departure and arrival time that is depicted from Fig. 14 to 17, thus, it can be determined the time slots of flight where most and least flights are delayed. Fig. 14 reveals that the largest number of flights are delayed during the period of 2 PM to 9 PM, however, the frequency is high because both delayed and not delayed flights are high. On the contrary, Fig. 15 discloses that the probability of flight delay soars during the period of 1 AM to 11 AM and 9 PM to 12 PM since in the winter season, there may be fog or visibility issues. On the other hand, Fig. 16 shows that the flight delay over arrival time rises linearly during the period of 6 AM to 10 PM since high frequency of air traffic over the time. In Fig. 17, with respect to the probability of the flight delay over arrival time, most of the delays are shown to occur from 6 AM to 2 PM.

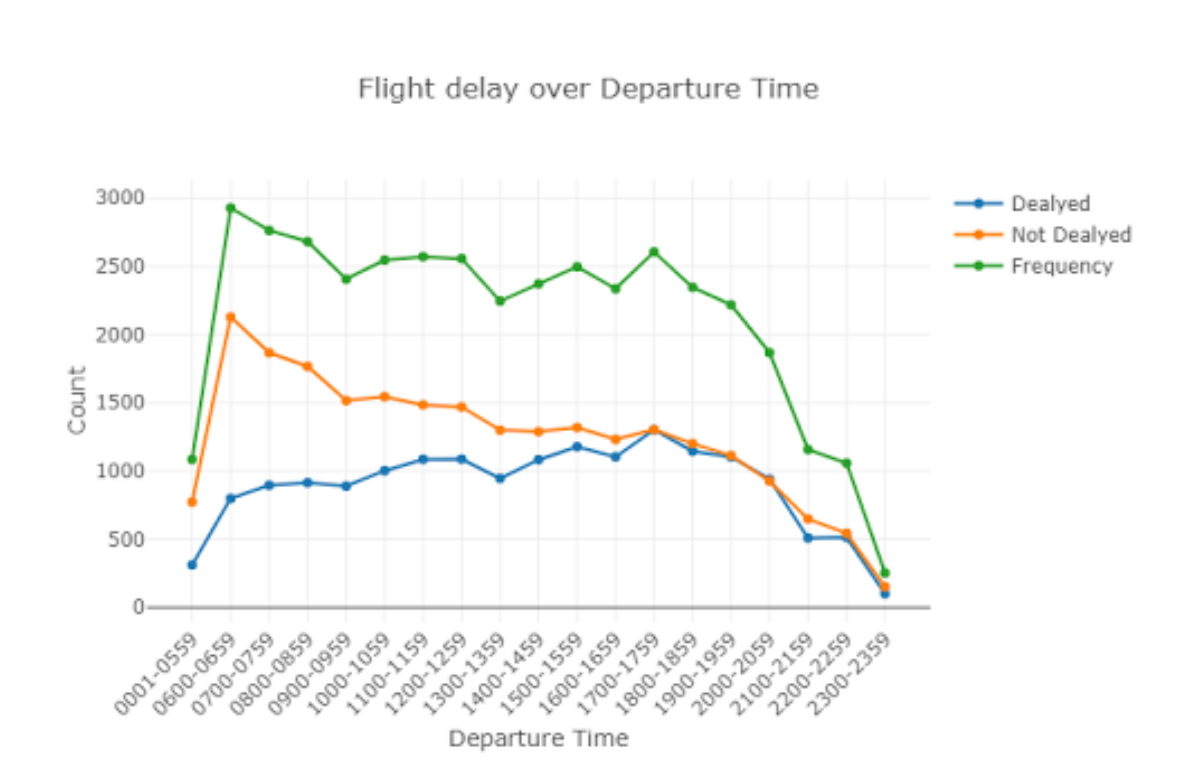

Fig. 14. The number of flights delayed over departure time of the flight

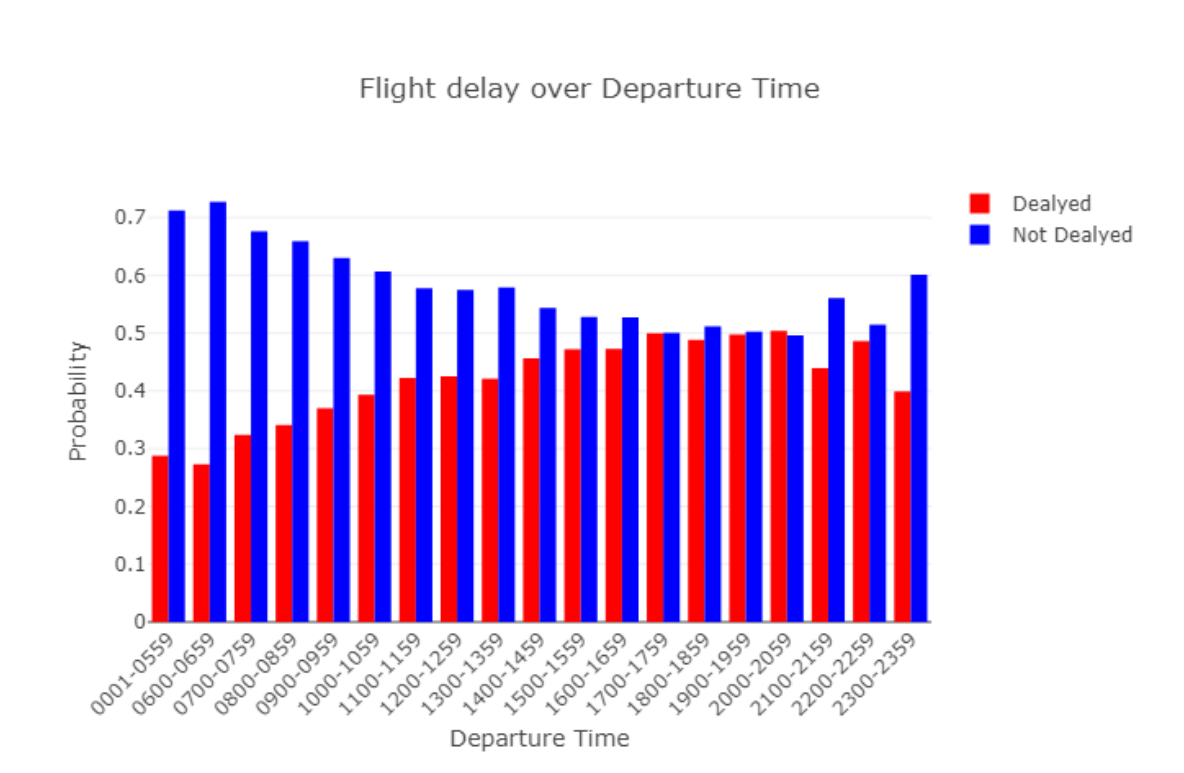

Fig. 15. The probability of flight delayed with respect to departure time of the flight

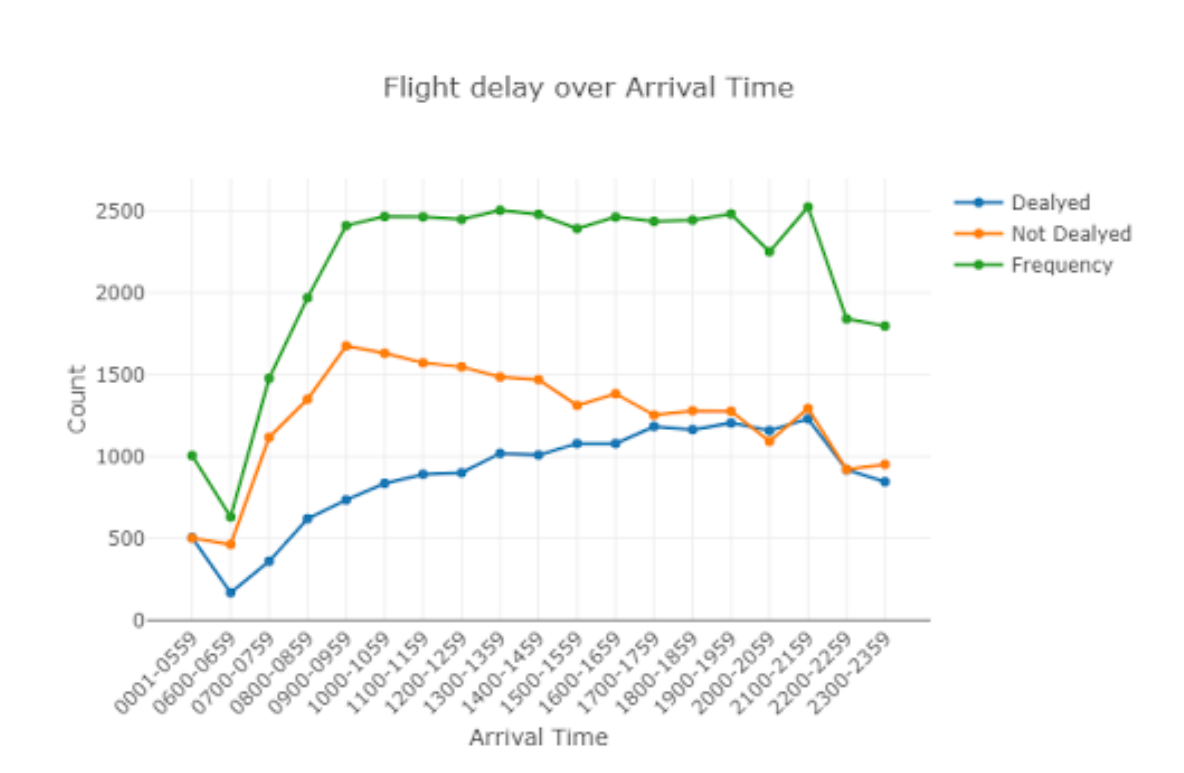

Fig. 16. The number of flights delayed over arrival time of the flight

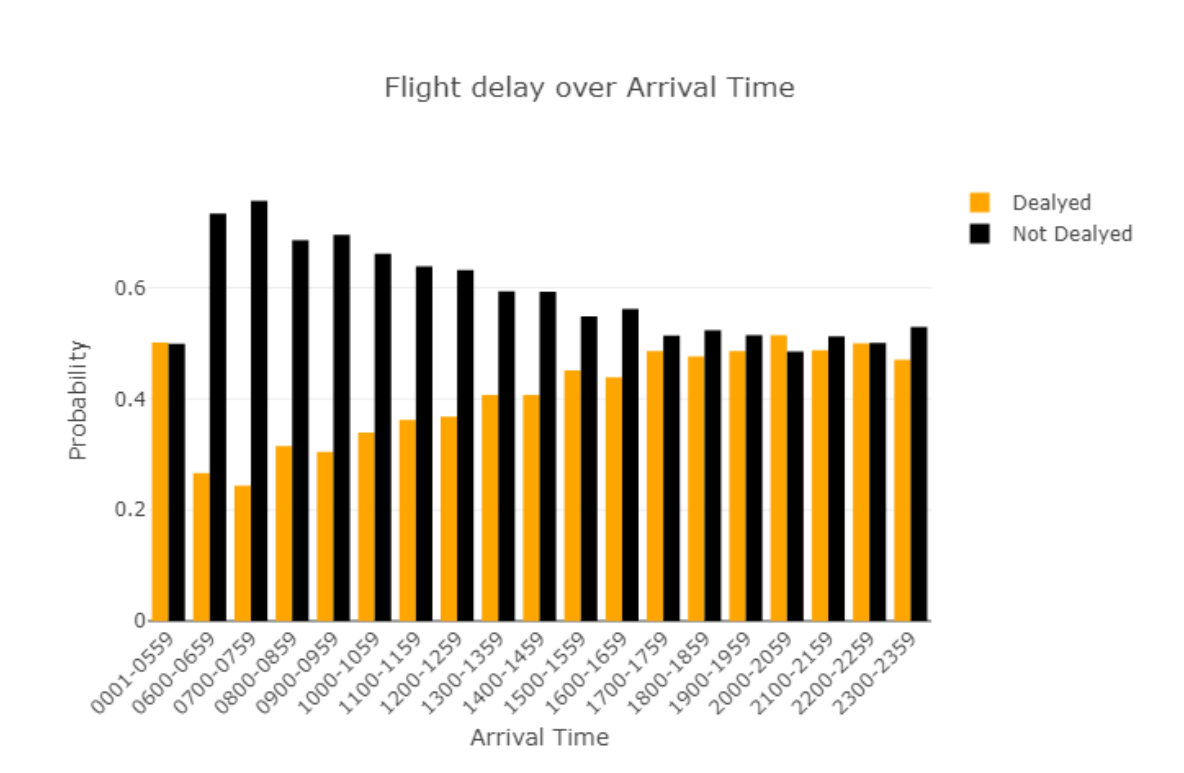

Fig. 17. The probability of flight delayed with respect to arrival time of the flight

## IV. Methodology

### *A. Fully Connected Neural Network*

The latest advancements in machine learning highlight the fully connected neural network as a well-established technique for representing data in high dimensions. Referred to as a deep belief network in literature, this architecture extends the traditional multiple layer perceptron by incorporating additional layers. Essentially, it comprises an input and output layer alongside multiple hidden layers, arranged sequentially. Data is initially fed into the input layer, followed by the placement of hidden layers, which play a crucial role in identifying patterns within the data for improved representation. Subsequently, the output layer is positioned to classify the transformed data emerging from the final hidden layer. The gradient-based backpropagation algorithm is utilized to optimize the model and minimize generalization error. Weight updates are based on a chosen loss function, with binary cross-entropy loss being employed in this study.

### *B. Random Forest*

Random forest (Breiman, 2001) is an ensemble of decision trees where each tree is built from a sample drawn from the training set. It uses feature randomness when building each single tree to try to generate an uncorrelated forest of trees whose classification is more precise than that of any particular tree. To give more randomness in building a random forest, some random subset of given features or all features are considered for best split, while splitting operations on each node (Ho, 1998). Size of the random subset is passed by the user as a hyper parameter. Individual decision trees suffer from high variance problems that lead to overfitting of the tree estimator. Random forest overcomes the problem of high variance in individual trees by providing above-mentioned two types of randomness. Random subset samples make different errors, and thus estimators generalize well by predicting the class having majority votes out of each predictor that facilitates in cancelling out the errors. However, random forests suffer a little bit from the increased bias problem, but variance needs to be taken care rather than bias.

### *C. XGBoost*

XGBoost (Chen & Guestrin, 2016) is an "Extreme Gradient Boosting" based on the concept of gradient boosting (Friedman, 2001) trees. The main feature that makes it different from other gradient boosting based techniques is its objective function, which consists of two parts, training loss and regularization term that can be seen in Eq. (1).

$$L(\phi) = \sum_{i} l(\hat{y}_i, y_i) + \sum_{k} \Omega(f_k) \tag{1}$$

$$Where,\ \Omega(f) = \gamma\tau + \frac{1}{2}\lambda\|w\|^2 \tag{2}$$

The training loss measures how predictive the model is with respect to the training data, while on the other side, the regularization term controls the complexity of the model that helps to generalize the model better. In case of XGBoost, the Taylor expansion of the loss function up to the second order is used to expand the loss function (polynomials). XGBoost is an optimal joint effort of hardware and software optimization techniques to yield valid results by conducting a smaller amount of computing resources in a little amount of time.

### *D. FCNN and Tree Based Ensemble Classifiers*

In this study classical machine learning as well as deep learning algorithms have been applied to solve the flight delay and cancellation problem. It has been observed that neither classical machine learning algorithms, nor deep learning algorithms suffice to achieve better accuracy on test data. Thus, a novel technique is proposed, which incorporates the advantage of both deep learning and classical machine learning algorithms. In classical machine learning, gradient learning based, and ensemble technique based advanced algorithms are used, namely, XGBoost and Random Forest. So far as a deep learning-based method is concerned, fully connected neural network (FCNN) is used with some advanced regularisation technique, which could be dropout (Srivastava et. al., 2014). Batch normalization (Ioffe and Szegedy, 2015) is also used to speed up the training process and to overcome the ill-conditioning of the Jacobian matrix and covariant shift problems. In novel techniques, first FCNN is trained on given data (Fig. 18) and then the output from the second last layer (Fig. 19) is saved as transformed training data. This approach is to increase the representation space of the original data. It is unlike the principal component analysis where dimensions are reduced. On the contrary, the dimension of original data is increased by passing it to a deep learning model for more comprehensive data representation. Here FC layers are treated as a feature extraction mechanism. The transformed data is passed to the tree-based ensemble classifiers inference. This solution is depicted in Fig. 20, where the proposed architecture is visualized. It is noticeable that data representation in this rich manner results in robust decision-making capability for classical machine learning algorithms.

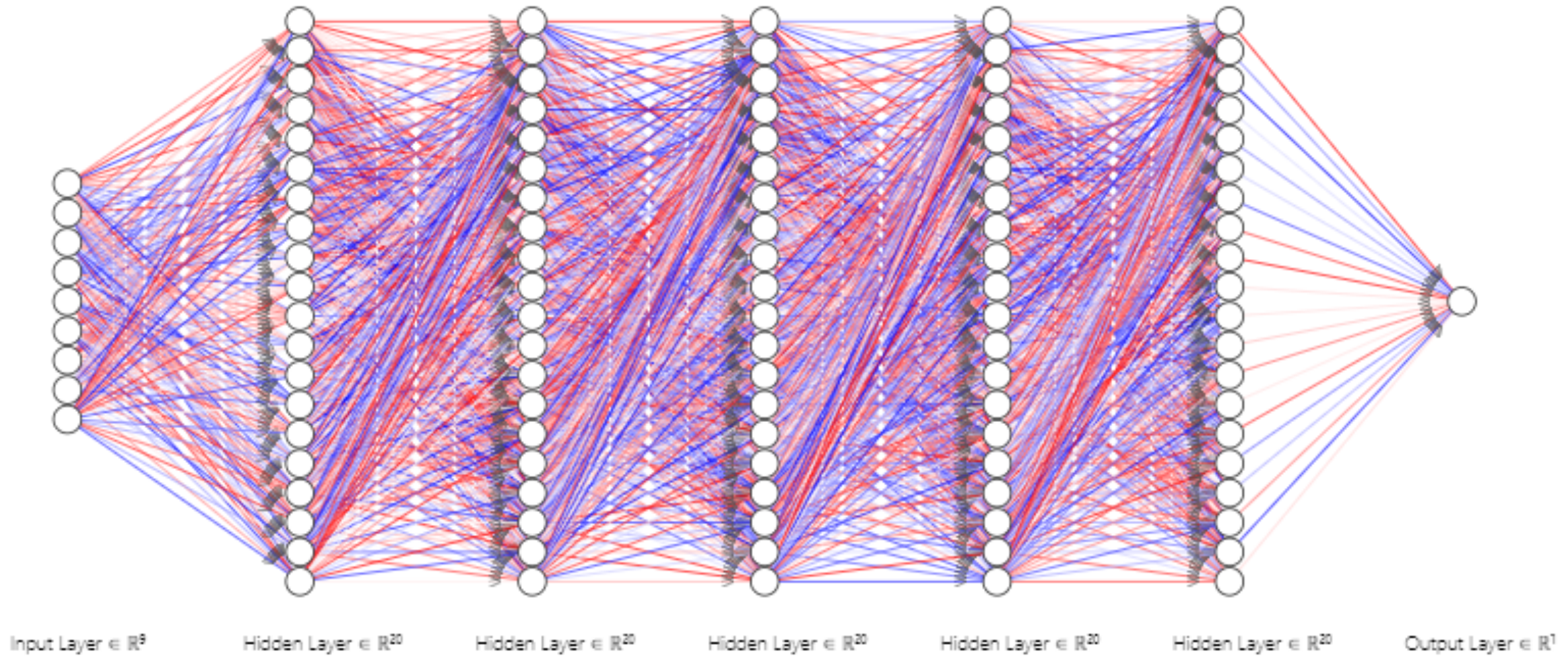

Fig 18: Illustration of fully connected neural network. First layer is input layer, whereas last layer is output layer. Layers in between are hidden layers.

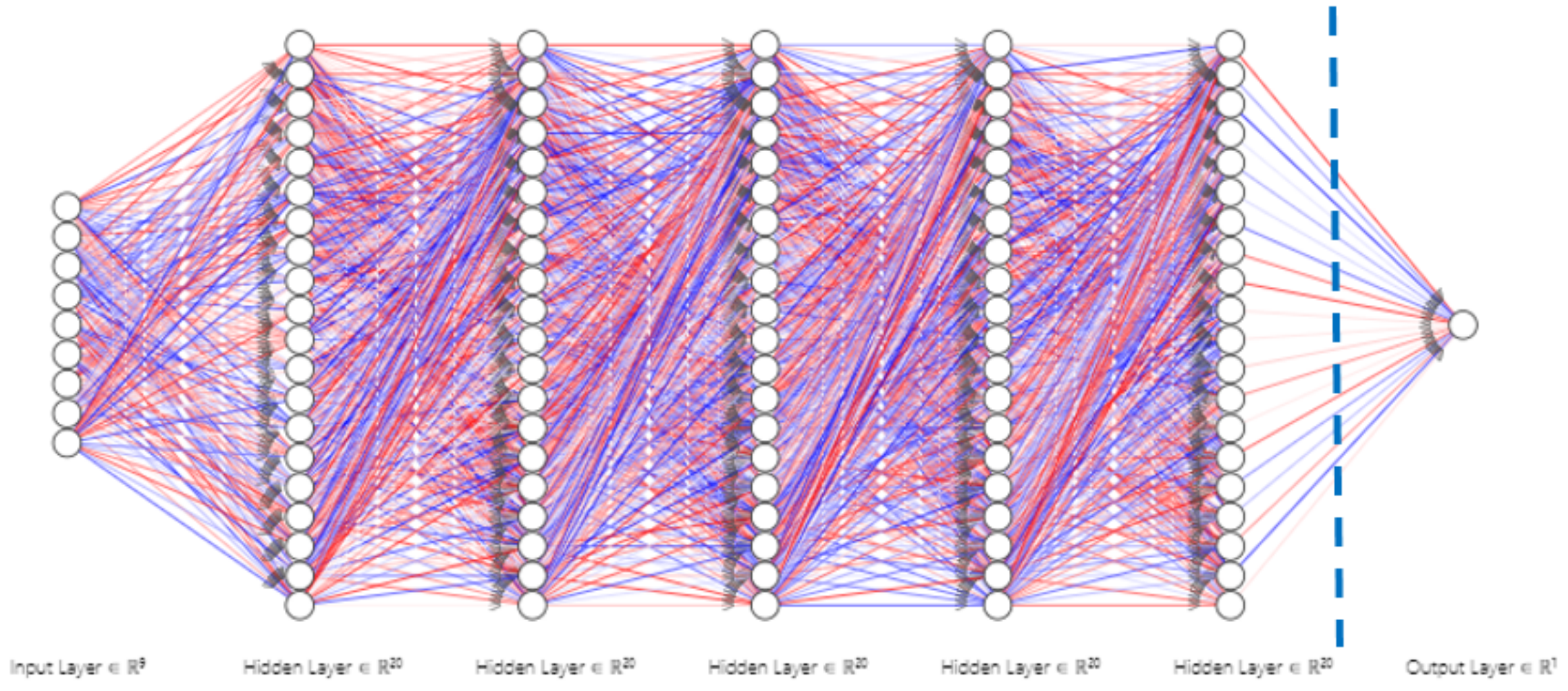

Fig 19: Output layer of FCNN is removed to get the activation of the second last layer.

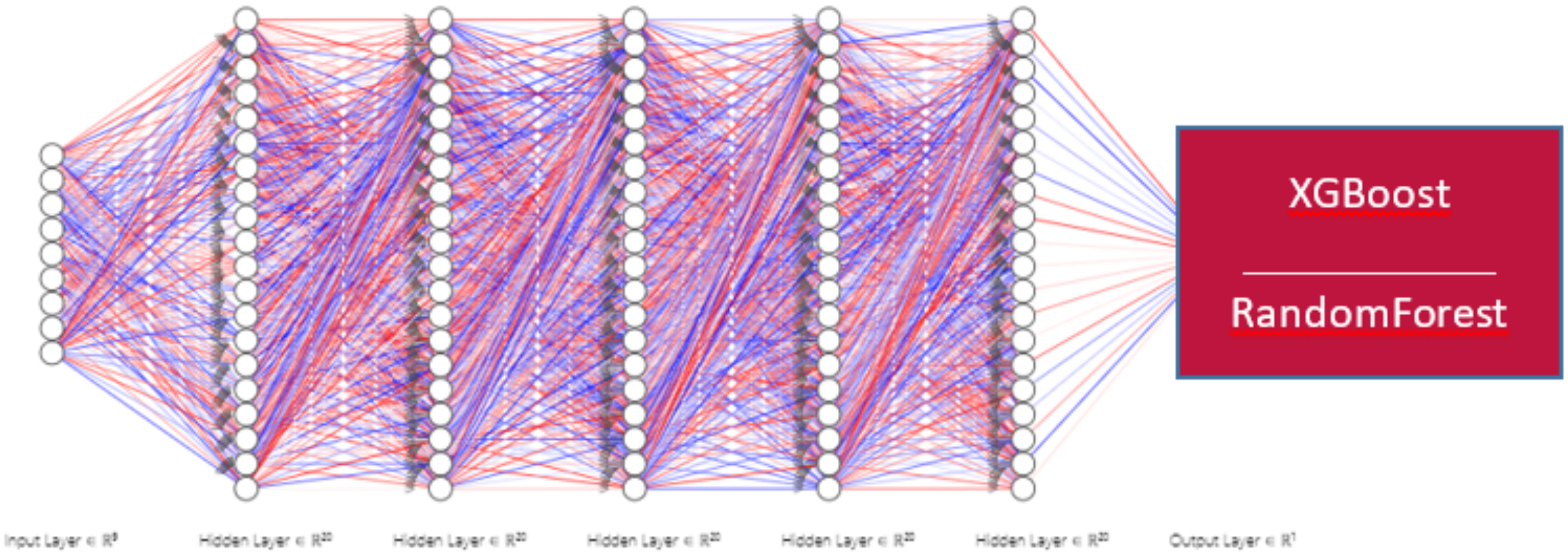

Fig 20: FCNN after detaching output layer and adding tree-based ensemble methods, which takes input as the activation of the Fig. 19 architecture. Output using tree-based ensemble methods is considered as final prediction.

## V. Empirical Results

In this section, a comprehensive analysis of above mentioned approaches for each of the three classification tasks on flight data is conducted.

**Table 2. Performances against each classifier.**

| Classifiers | Metric | FCNN + RDF | RDF | XGBoost | FCNN |
|---|---|---|---|---|---|
| **Departure Delay** | Accuracy | 90.57 | 85.6 | 91.6 | 88.6 |
| | Precision | 92.1 | 90.1 | 97.4 | 91.2 |
| | Recall | 88 | 79.6 | 85.5 | 84.8 |
| | f1-score | 90.5 | 84.4 | 91.1 | 88.2 |
| **Arrival Delay** | Accuracy | 93.4 | 86.6 | 94.38 | 92.2 |
| | Precision | 93.6 | 88 | 97.6 | 93.3 |
| | Recall | 93.1 | 84.3 | 91 | 92 |
| | f1-score | 93.4 | 86.1 | 94.2 | 92.2 |
| **Total Delay** | Accuracy | 92.5 | 83.1 | 84.5 | 88.7 |
| | Precision | 91.5 | 83.05 | 85.6 | 83 |
| | Recall | 91.4 | 83.05 | 85.15 | 94.4 |
| | f1-score | 91.4 | 83 | 85.15 | 88.3 |

### *A. Departure and Arrival Delay Analysis*

In this task, XGBoost shows significantly better performance over other classifiers for accuracy, F1, precision, while in terms of recall, FCNN + RDF performs better among the used classifiers. Table 2 presents each metric value against each classifier. In addition, performance of Arrival Delay classification is dominated by the XGBOOST classifier. However, FCNN + RDF is performing better in recall, nonetheless XGBoost performs well for the rest of other metrics.

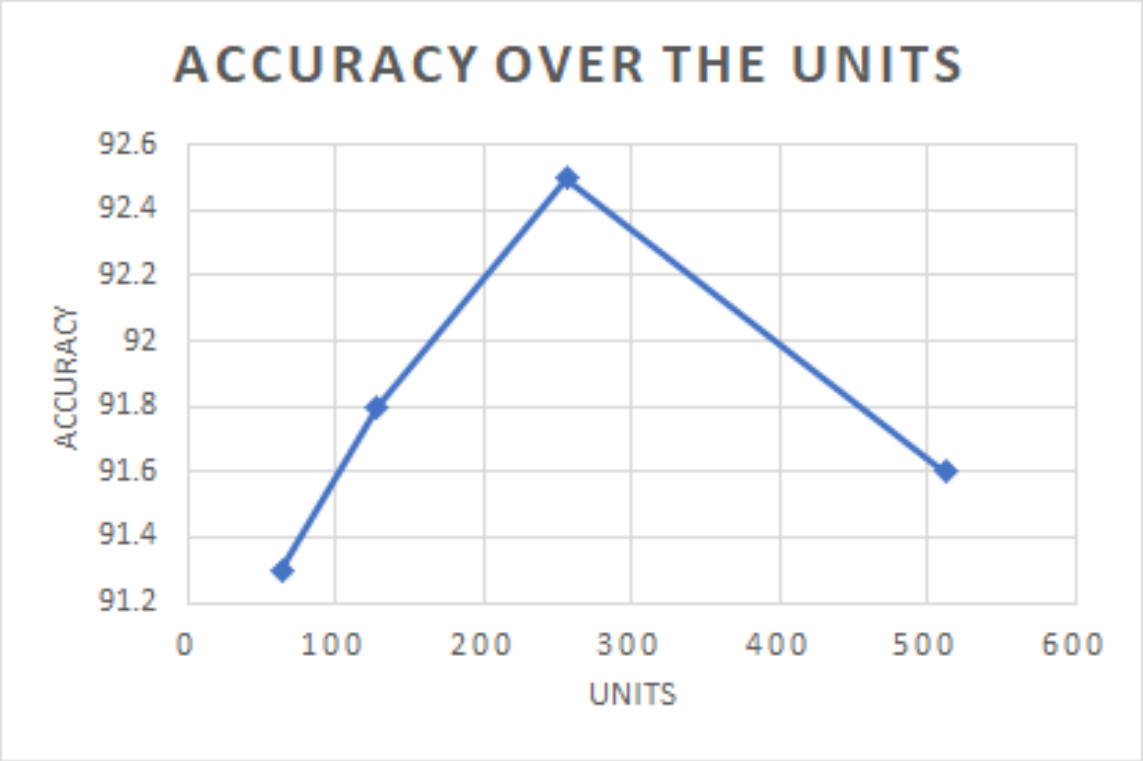

Fig 21: For the second last layer of Fig. 18 different number of hidden units have been tried. Plot demonstrates that 256 hidden units shows better results in terms of accuracy.

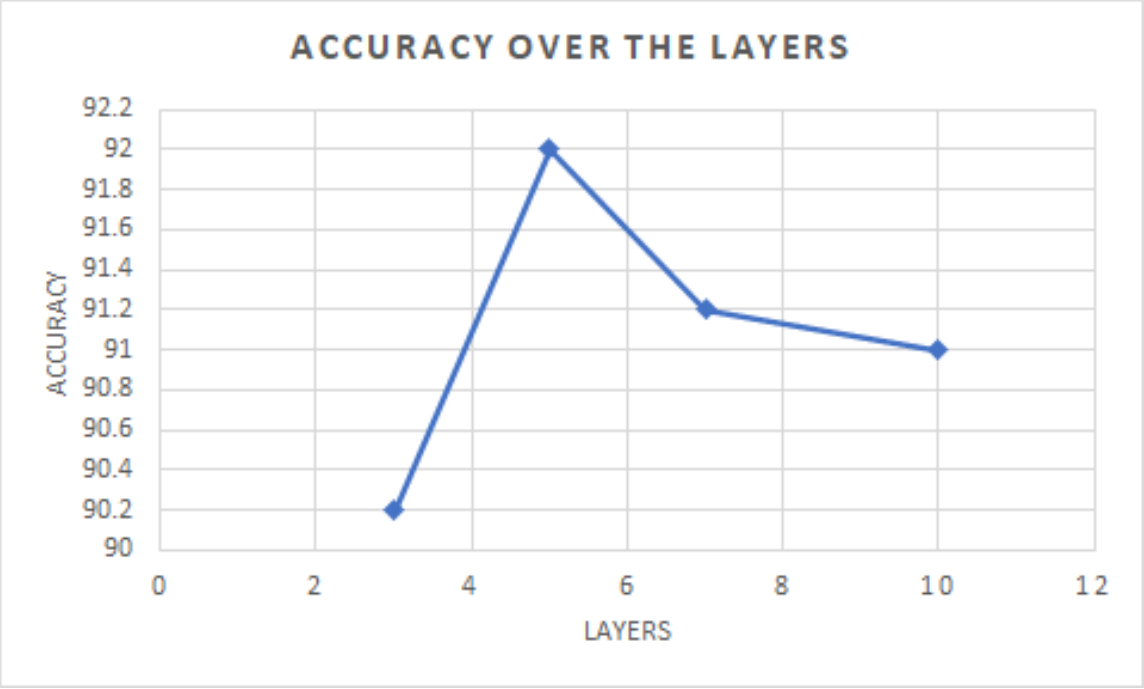

Fig 22: Experiment has been conducted for variable layers sizes. Plot demonstrates that beyond 5 hidden layers models give diminishing returns over accuracy if layer size is increased.

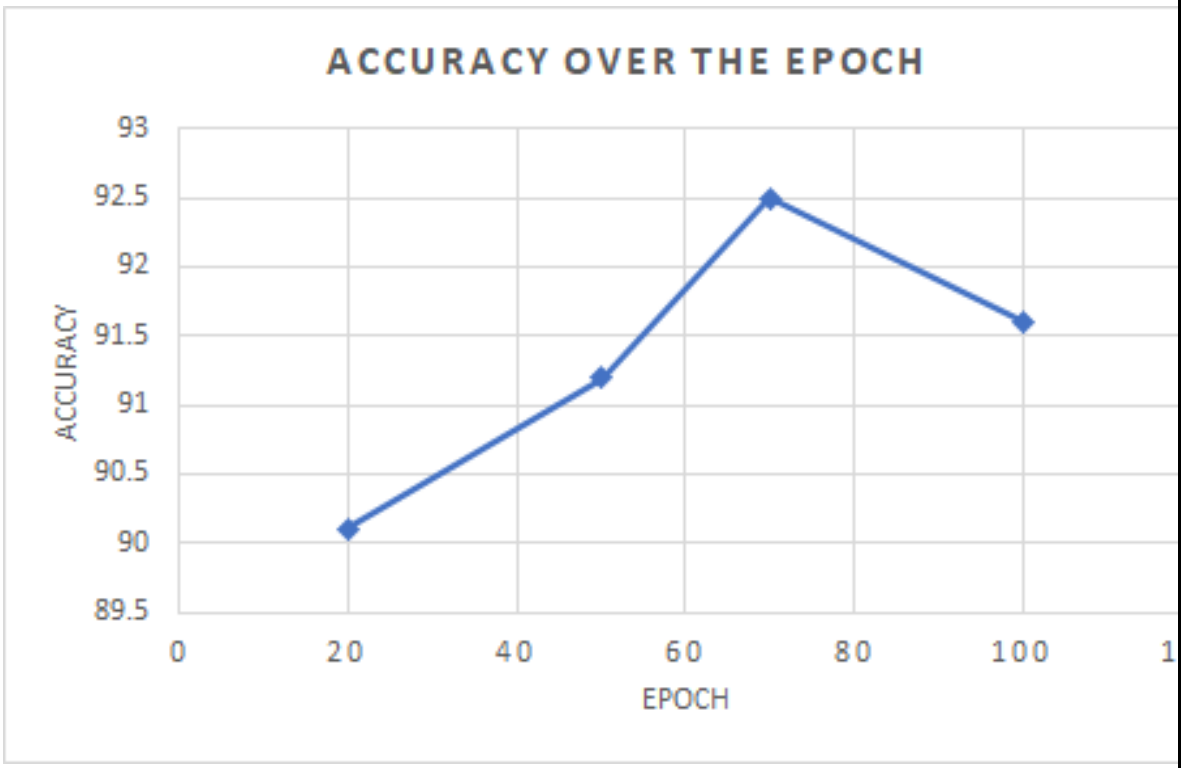

Fig 23: Accuracy over epoch

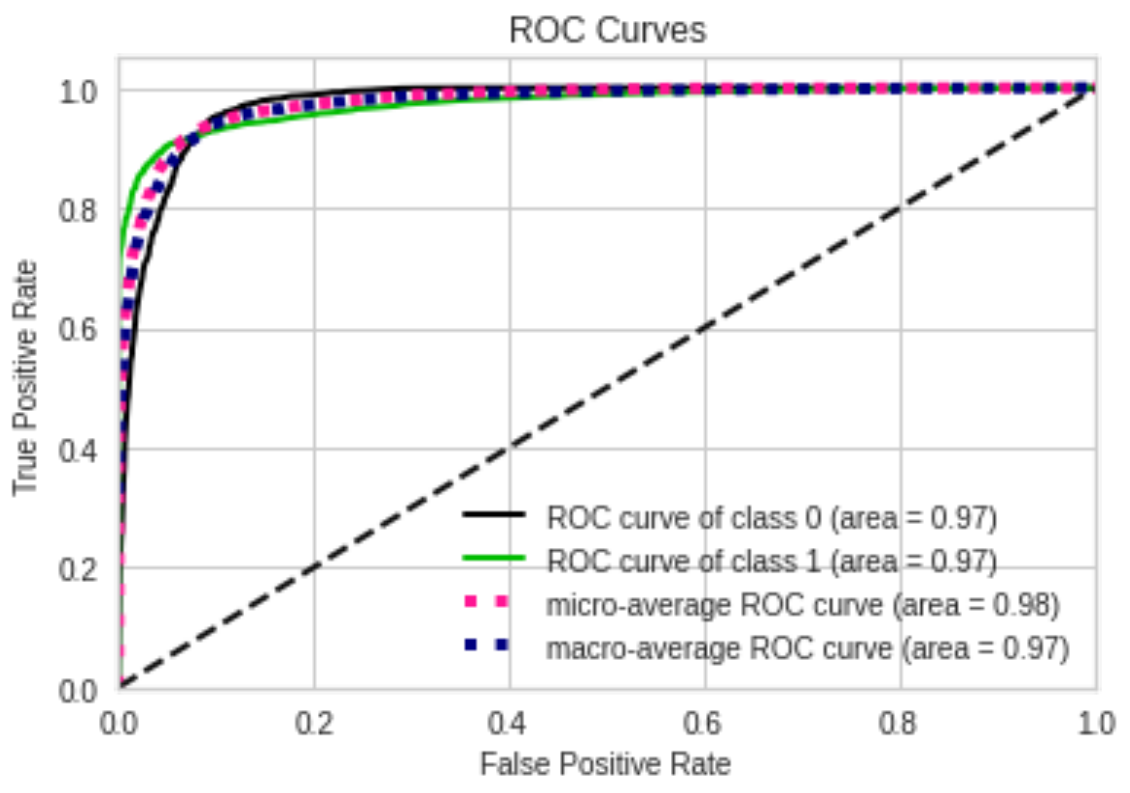

Fig 24: Total delay ROC curve of FCNN + RDF

### *B. Total Delay Analysis*

In total delay classification, FCNN + RDF performs exceptionally well for all the metrics, except recall, where FCNN performs better. There is a reasonable gap between the performance of FCNN + RDF and other used classifiers. In this paper, the novel approach of FCNN + RDF shows ground-breaking results for total delay task. Fig. 24 shows that the AUC value is 0.97 for both classes (i.e., class 0 and class 1), which is a decent value in terms of the quality of a classifier. Pertaining to FCNN+RDF, several experiments have been conducted, on the number of hidden units in the second last layer, number of hidden layers, and of epochs size. Plots corresponding to this experiment are shown in Fig. 21. It can be noticed that the size of hidden units should be 250 to produce optimal results. This value should be neither very high nor very low. The same observation with the number of hidden layers (the value is five in this proposed approach) shows better results, while for less or more hidden layer values, the model does not perform well. The novel approach of FCNN + RDF helps in achieving significantly better result than the traditional ensemble-based tree methods. One major reason for the observed enhanced performance is the rich data representation of original data via higher dimensional data. The original data representation using predefined numbers of features are not enough to capture the concept in data, whereas FCNN helps in learning the concepts present in data in a reasonably well manner. Once data is properly represented using higher dimensions, the ensembled based tree classifiers are applied to classify the newly transformed rich data representation. Both XGBoost and Random Forest have been tried, although for new data the Random Forest produces better results on defined metric measures.

### *C. Accuracy over Units, Layers, Epochs, and Classifiers*

This subsection discusses the impact of hidden units, hidden layers, and epochs size on learning. Fig. 21 makes it apparently visible, as the model moves beyond 250 hidden units, that the accuracy keeps decreasing. But on contrary, as the number of hidden units decreases, the accuracy begins to show lower quality. This behaviour advocates that neither very high dimensional representation, nor low dimensional representation help in achieving significant performance. For the studied model, representing data in 250 dimensional features helps in attaining remarkable results. For the number of hidden layers, Fig. 22 shows that a greater number of hidden layers make the model more complex, which results in in overfitting the model. Conversely, a lower number of hidden layers are incapable of fitting the data itself, thus causing high generalization error. Via experiment, it has been concluded that, five hidden layers, with 250 hidden units fits the data well for the total delay classification task. It is a known fact that higher epochs sometimes result in overtrained weights, which produces a high generalization error. Fig. 23 outlines that an epoch size of 75 works well for out data and proposed network architecture. In addition, the accuracy over each model with respect to each classifier are shown in Fig. 25.

### *D. ROC Curves of All Models*

Receiver Operator Characteristic (ROC) curve is the key tool to measure the quality of the classifier. This subsection includes a detailed discussion on the above-mentioned classifier's characteristics. For further discussion, this is referred to the ROC curve of the best classifiers of each task only. For all three ROC curves in Fig. 24, 26, and 27, there are no notable difference in the AUC values. However, the shape of the curve reveals some meaningful difference in classifier's characteristics. ROC curve of XGBoost for arrival delay task suggests that XGBoost separates data in positive and negative classes in a better way. Until the true positive rate of 0.8, it has 0.0 false positive rate. This behaviour shows that XGBoost is an excellent classifier. For the other two tasks, namely departure delay and total delay, the classifier performs better for class 1 (delay or not delay) as compared to class 0, although, it also performs quite well for class 0. Overall, each of the three classifier's performs equally well for class 1, whilst, for class 0 their performances differ by some extent.

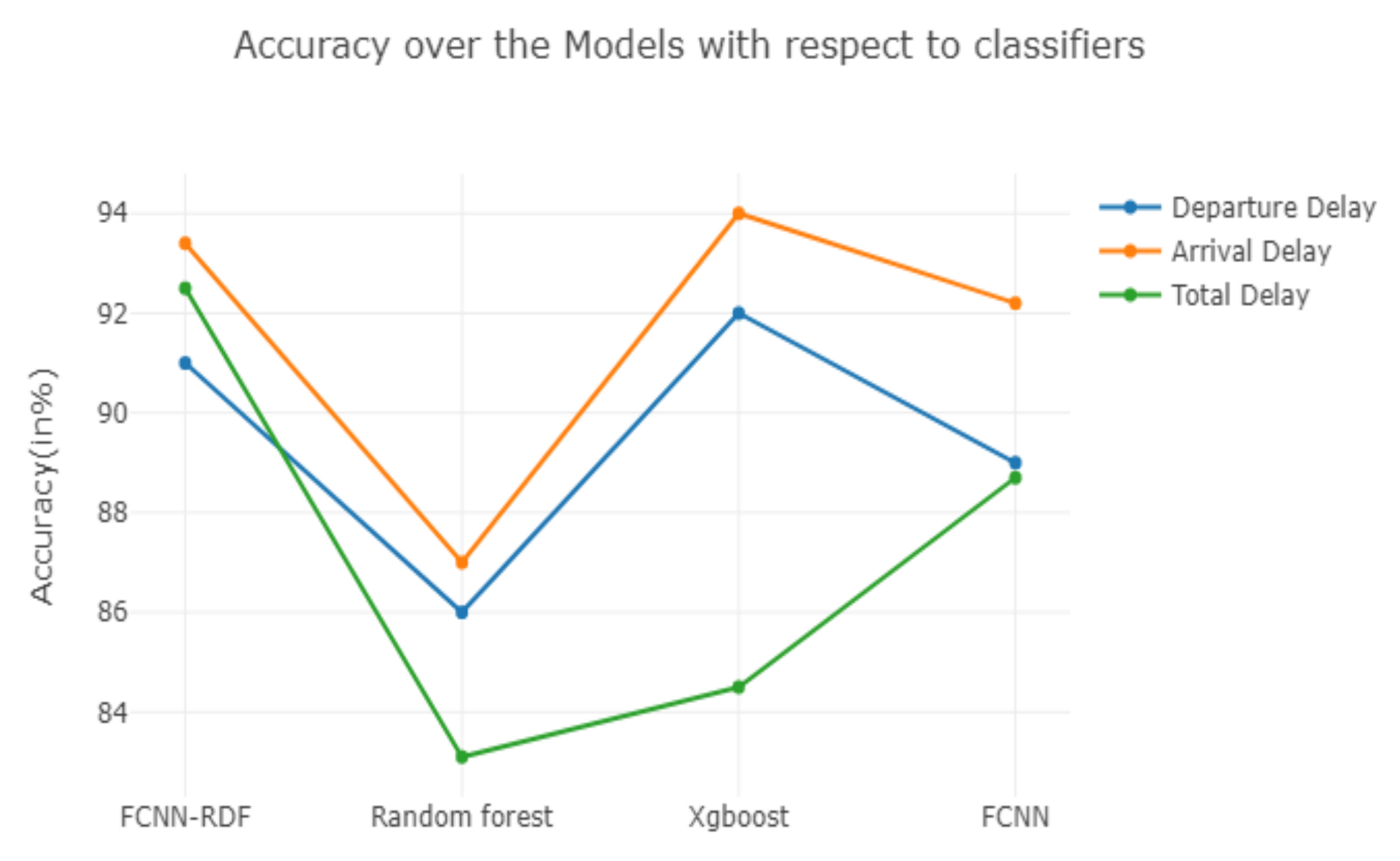

Fig. 25. Accuracy over the models with respect to classifiers

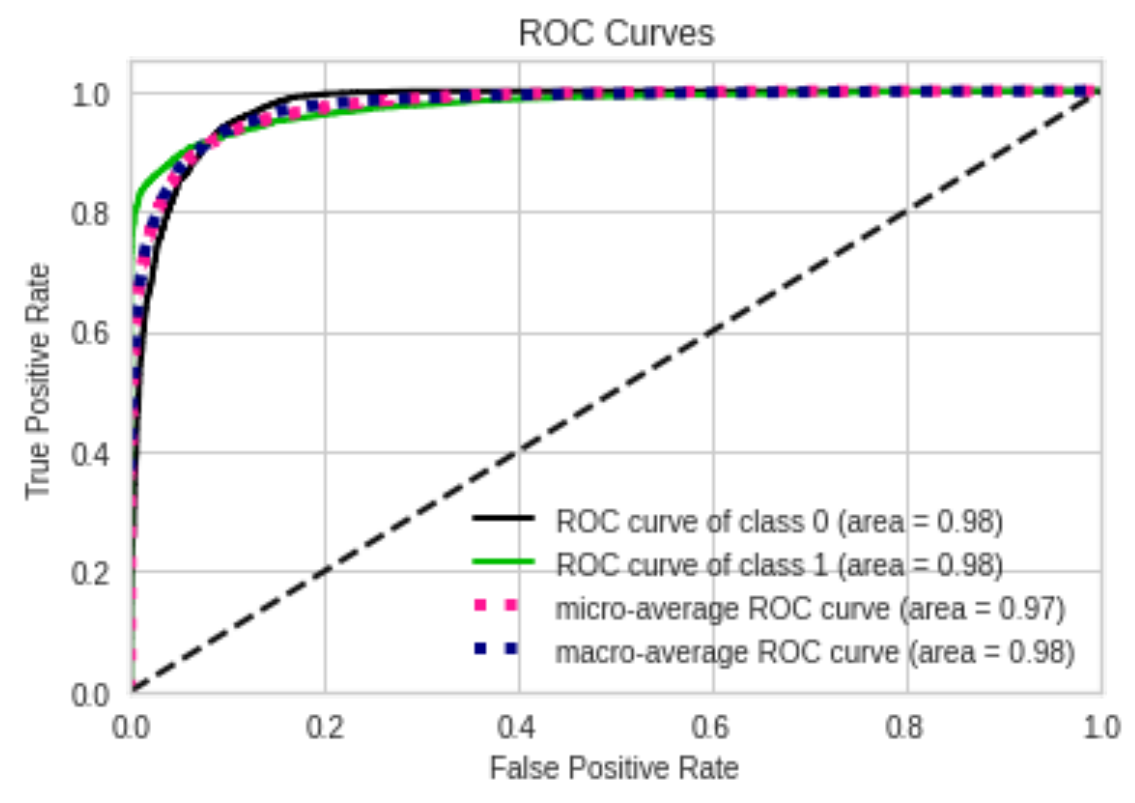

Fig 26: XGBoost ROC curve for departure delay

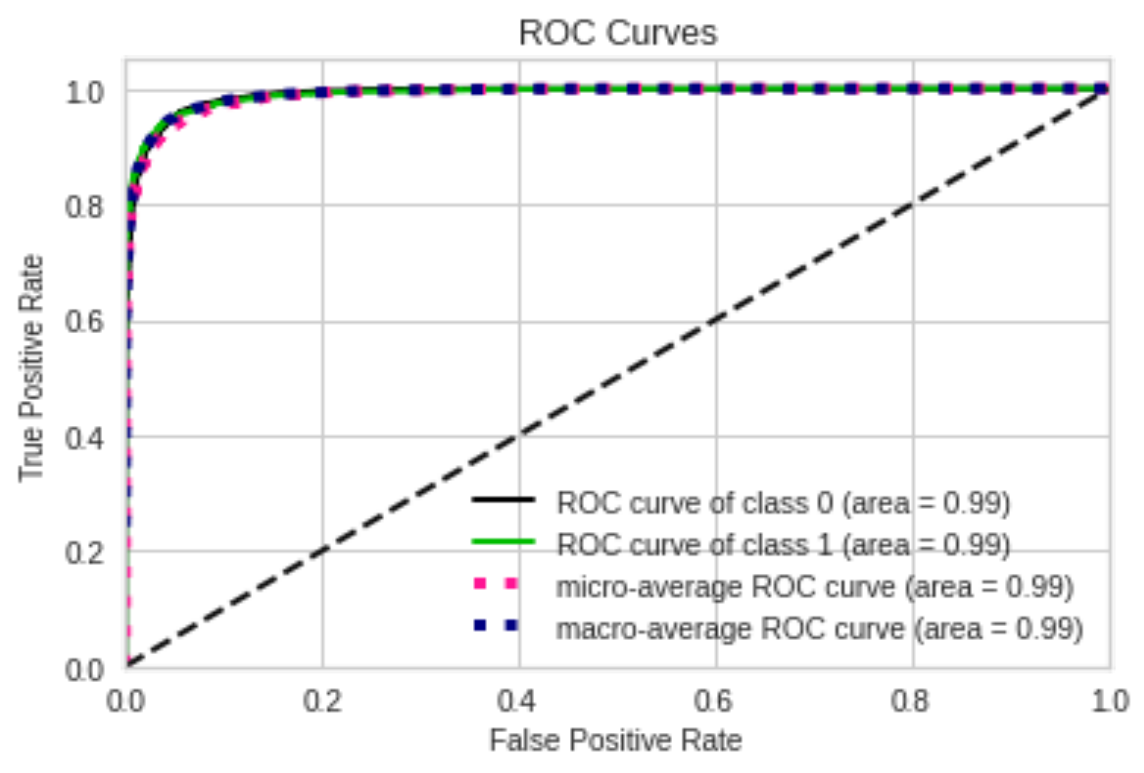

Fig 27: XGBoost ROC curve for arrival delay

## VI. Discussion:

The study of flight delays in the context of U.S. airlines shows critical insights into the challenges faced by the aviation industry to manage and mitigate delays. The hybrid approach proposed for flight delays in this research, which combines deep learning with classical machine learning techniques, demonstrates significant improvements to predict flight delays compared to traditional methods. The effectiveness of this work is expressed by handling complex and large-scale datasets. The higher accuracy, precision, and F1-score, in the case of departure delay, arrival delay, and total delay classification, are considerable.

The analysis of flight delay patterns across different airlines, states, and times intervals provides valuable insights for operational decision-making. For instance, understanding that delays are more frequent during peak travel seasons and specific times of the day helps airlines optimize their schedules and resource allocation. The identification of high-risk periods for delays, such as early morning and late evening flights, could lead to targeted interventions to reduce delays, such as adjusting flight schedules or enhancing ground support during these times. The departure delay problem offers solutions that can be helpful to airlines to schedule the flight, and crew members properly which can lessen the cost significantly. Airlines can also contact passengers and update them continuously regarding their flights. In this case, if the passengers are in known of their flight status, then they can manage their next schedule appropriately. Overall, this approach can reduce the anxiety level of passengers as well as lower their expenses, since suddenly changes in schedule can add to their trip cost. Similarly, the analysis applies for the flight arrival delay problem. In the last model, flight cancellations, if passengers are aware of cancellations that might happen then they can make decisions for the second option, so if flight cancellation occurred then the second option can be chosen.

## VII. Conclusion and Future Scope:

In conclusion, this study presents a novel hybrid approach that effectively addresses the flight delay classification problem by leveraging the strengths of both deep learning and classical machine learning techniques. The proposed method not only improves predictive performance but also provides actionable insights that can be used to optimize airline operations and reduce the negative impacts of flight delays on both airlines and passengers.

The future scope of this research could involve exploring more advanced deep learning architectures, such as recurrent neural networks (RNNs) or transformers, which may further enhance the ability to capture temporal patterns in flight delay data. Additionally, integrating real-time data sources, such as weather updates and air traffic control communications, could lead to more dynamic and responsive delay prediction models. Moreover, expanding the dataset to include global flight data and incorporating factors like international airspace regulations and cross-border operations could make the model applicable to a broader range of airlines. This would be particularly valuable for airlines operating in regions with diverse operational challenges. Finally, developing a user-friendly decision support system based on the proposed model could provide airlines with a practical tool to anticipate and manage delays more effectively, ultimately improving overall efficiency and customer satisfaction.

**References:**

1. World Bank, 2013. Air Transport, Passenger Carried. http://data.worldbank.org/indicator/IS.AIR. PSGR (accessed 03.17.13).
2. Ball, M., Barnhart, C., Dresner, M., Hansen, M. Neels, K., Odoni, A., Peterson, E., Sherry, L., Trani, A., Zou, B., 2010. Total Delay Impact Study. NEXTOR Final Report Prepared for the U.S. Federal Aviation Administration. http://www.nextor.org/pubs/TDI_Report_Final_11_03_10.pdf.
3. Joint Economic Committee (JEC), 2008. Your Flight has been Delayed Again: Flight Delays Cost Passengers, Airlines, and the US Economy Billions. http:// jec.senate.gov/index.cfm?FuseAction=Reports.Reports&ContentRecord_id=11116dd 7-973c-61e2-4874a6a18790a81b&Region_id=&Issue_id (accessed 03.14.13)
4. Cook, A., Tanner, G., Anderson, S., 2004. Evaluating the True Cost to Airlines of One Minute of Airborne or Ground Delay. Report Prepared by the University of Westminster. Performance Review Unit, Eurocontrol.
5. Zou, Li, Chen, Xueqian, 2017. The effect of code-sharing alliances on airline profitability. J. Air Transport. Manag. 58, 50–57
6. Boeing, 2011. Current Market Outlook: 2011–2030. http://www.boeing.com/commercial/cmo/index.html (accessed 03.12.12)
7. Yimga, Jules, Gorjidooz, Javad, 2019. Airline schedule padding and consumer choice behavior. J. Air Transport. Manag. 78, 71–79.
8. Forbes, Silke J., Lederman, Mara, Wither, Michael J., 2019. Quality disclosure when firms set their own quality targets. Int. J. Ind. Organ. 62, 228–250.
9. Zou, B., & Hansen, M. (2014). Flight delay impact on airfare and flight frequency: A comprehensive assessment. *Transportation research part E: logistics and transportation review,* 69, 54-74.
10. Lambelho, M., Mitici, M., Pickup, S., & Marsden, A. (2020). Assessing strategic flight schedules at an airport using machine learning-based flight delay and cancellation predictions. *Journal of Air Transport Management*, *82*(October 2019). https://doi.org/10.1016/j.jairtraman.2019.101737
11. Yu, B., Guo, Z., Asian, S., Wang, H., & Chen, G. (2019). Flight delay prediction for commercial air transport: A deep learning approach. *Transportation Research Part E: Logistics and Transportation Review*, *125*(October 2018), 203–221. https://doi.org/10.1016/j.tre.2019.03.013
12. Yimga, J. (2020). Journal of Air Transport Management Price and marginal cost effects of on-time performance : Evidence from the US airline industry ☆. *Journal of Air Transport Management*, *84*(November 2019), 101769. https://doi.org/10.1016/j.jairtraman.2020.101769
13. Rebollo, J. J., & Balakrishnan, H. (2014). Characterization and prediction of air traffic delays. *Transportation Research Part C: Emerging Technologies*, *44*, 231–241. https://doi.org/10.1016/j.trc.2014.04.007
14. Liu, Y., Liu, Y., Hansen, M., Pozdnukhov, A., & Zhang, D. (2019). Using machine learning to analyze air traffic management actions: Ground delay program case study. *Transportation Research Part E: Logistics and Transportation Review*, *131*(December 2018), 80–95. https://doi.org/10.1016/j.tre.2019.09.012
15. Breiman, L. (2001). Random forests. *Machine learning*, 45(1), 5-32.
16. Ho, T. K. (1998). The random subspace method for constructing decision forests. *IEEE transactions on pattern analysis and machine intelligence*, 20(8), 832-844.

17. Chen, T., & Guestrin, C. (2016, August). Xgboost: A scalable tree boosting system. *In Proceedings of the 22nd acm sigkdd international conference on knowledge discovery and data mining* (pp. 785-794).
18. Friedman, J. H. (2001). Greedy function approximation: a gradient boosting machine. *Annals of statistics*, 1189-1232.
19. Srivastava, N., Hinton, G., Krizhevsky, A., Sutskever, I., & Salakhutdinov, R. (2014). Dropout: a simple way to prevent neural networks from overfitting. *The journal of machine learning research*, 15(1), 1929-1958.
20. Ioffe, S., & Szegedy, C. (2015, June). Batch normalization: Accelerating deep network training by reducing internal covariate shift. *In International conference on machine learning* (pp. 448-456). PMLR.